\documentclass{article} 
\usepackage{iclr2027_conference,times}

\usepackage[utf8]{inputenc}
\usepackage[T1]{fontenc}
\usepackage{booktabs}
\usepackage{enumitem}
\usepackage{amsfonts}
\usepackage{amsmath}
\usepackage{amssymb}
\usepackage{mathtools}
\usepackage{algorithm}
\usepackage{algpseudocode}
\usepackage{graphicx}
\usepackage{subcaption}
\usepackage{wrapfig}
\usepackage{microtype}
\usepackage{fvextra}
\usepackage{xcolor}
\usepackage{array}
\usepackage{colortbl}
\usepackage{placeins}
\usepackage{hyperref}
\usepackage{url}
\usepackage[capitalise]{cleveref}
\usepackage{fontawesome5}

\makeatletter
\@ifundefined{theHALG@line}
  {\newcommand{\theHALG@line}{\thealgorithm.\arabic{ALG@line}}}
  {\renewcommand{\theHALG@line}{\thealgorithm.\arabic{ALG@line}}}
\makeatother

\let\cite\citep

\newcommand{\tablestyle}[2]{\setlength{\tabcolsep}{#1}
                            \renewcommand{\arraystretch}{#2}
                            \centering
                            \footnotesize}
\newcolumntype{x}[1]{>{\centering\arraybackslash}p{#1pt}}
\newcolumntype{y}[1]{>{\raggedright\arraybackslash}p{#1pt}}
\newcolumntype{z}[1]{>{\raggedleft\arraybackslash}p{#1pt}}
\definecolor{graycolor}{gray}{.9}
\definecolor{bestcolor}{HTML}{e7f2fc}
\definecolor{secondcolor}{HTML}{ffe5ec}

\newcommand{\best}[1]{\cellcolor{bestcolor}{#1}}
\newcommand{\second}[1]{\cellcolor{secondcolor}{#1}}

\definecolor{pagelinkcolor}{HTML}{1A5FB4}
\newcommand{\projectpageurl}{https://plurato.github.io/OctLLM-page/}
\newcommand{\projectpage}{%
  {\normalfont\color{pagelinkcolor}\hypersetup{pdfborder={0 0 0}}%
   \faGlobe\hspace{0.35em}%
   \textbf{Project Page}:\hspace{0.4em}%
   \href{\projectpageurl}{\fontfamily{zi4}\selectfont\projectpageurl}}}

\DefineVerbatimEnvironment{PromptBox}{Verbatim}{
  breaklines=true,
  breaksymbolleft={},
  fontsize=\footnotesize,
  frame=single,
  framerule=0.4pt,
  framesep=6pt
}
\newsavebox{\exampleboxsave}
\newenvironment{ExampleBox}
  {\par\addvspace{0.5\baselineskip}%
   \noindent
   \setlength{\fboxrule}{0.4pt}%
   \setlength{\fboxsep}{6pt}%
   \begin{lrbox}{\exampleboxsave}%
   \begin{minipage}{\dimexpr\linewidth-2\fboxsep-2\fboxrule\relax}%
   \footnotesize}
  {\end{minipage}%
   \end{lrbox}%
   \fbox{\usebox{\exampleboxsave}}%
   \par\addvspace{0.5\baselineskip}}

\title{Octrees as an Explicit 3D Language}

\author{%
  Ran Dan\textsuperscript{1}\hspace{0.75em}%
  Si-Tong Wei\textsuperscript{1}\hspace{0.75em}%
  Pengfei Xiong\textsuperscript{2}\hspace{0.75em}%
  Wei Zhang\textsuperscript{2}\hspace{0.75em}%
  Yadong Mu\textsuperscript{1}\hspace{0.75em}%
  Peng-Shuai Wang\textsuperscript{1}\addtocounter{footnote}{1}\thanks{Corresponding author}\\[0.5em]
  \textsuperscript{1}Peking University\qquad
  \textsuperscript{2}Independent Researcher\\[0.45em]
  \projectpage
}

\iclrfinalcopy 
\begin{document}

\maketitle

\begin{figure}[h]
  \vspace{-14pt}
  \centering
  \includegraphics[width=\linewidth]{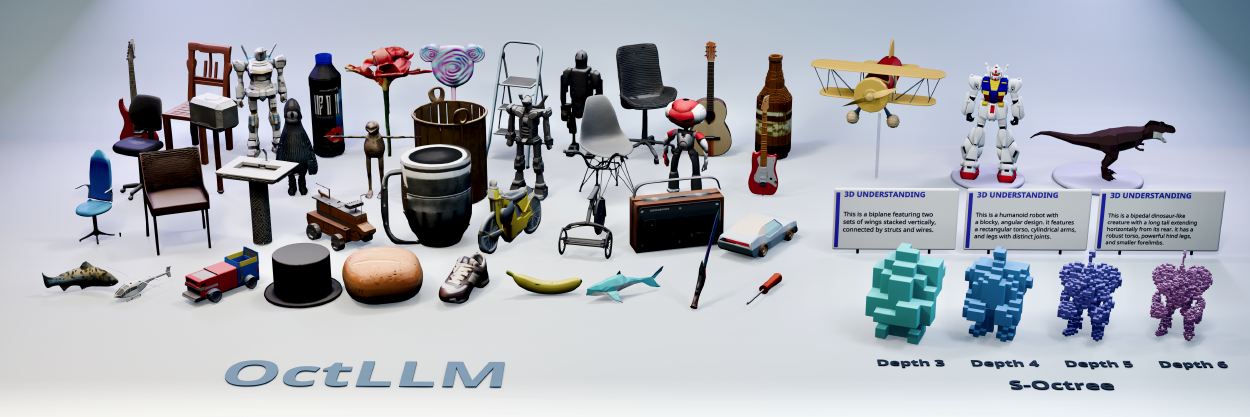}
  \caption{\textbf{OctLLM unifies 3D generation and 3D understanding in a single LLM.} Left: 3D assets generated by OctLLM from text and image conditions. Right: natural-language descriptions it produces for 3D inputs. Bottom right: a shape written as an S-Octree, the coarse-to-fine occupancy sequence OctLLM reads and writes, shown at octree depths 3 to 6. Generation and understanding share this explicit 3D language. Zoom in for a better view.}
  \label{fig:teaser}
\end{figure}

\begin{abstract}
Existing 3D large language models (LLMs) compromise on two fronts: they compress shapes into latent codebook indices or coordinate text, which removes spatial structure from what the model observes, and they acquire the 3D modality by fine-tuning the backbone, which overwrites its general language ability. We present OctLLM, which addresses both limitations. Geometry enters as an explicit 3D sequence of octree occupancy tokens. However, full octree sequences grow rapidly with depth; OctLLM therefore randomly empties penultimate-level nodes and omits descendants while preserving shape, yielding a shorter coordinate- and depth-anchored \emph{Sparse Octree} (S-Octree) for position-aware mask-modeling generation and 3D understanding. On the other front, existing methods introduce a new modality with full fine-tuning or LoRA, but full fine-tuning is costly, LoRA limits 3D capacity, and both modify the language pathway. OctLLM instead adds 3D capacity in parameters separate from the pretrained ones: mesh tokens are routed through independent trainable branches in a subset of blocks while text and image tokens retain the frozen vision-language pathway, and the two streams interact through shared self-attention. It trains far fewer parameters than full fine-tuning, yet sets a new state of the art among unified multimodal LLMs, lowering image-to-3D FID by $17.4\%$ and raising render-grounded captioning by $28.7$ points over ShapeLLM-Omni, while matching the backbone on general language benchmarks.
\end{abstract}

\section{Introduction} \label{sec:intro}

Large language models (LLMs) have advanced rapidly, with recent systems showing strong general-purpose reasoning and instruction-following ability across both proprietary and open settings~\cite{gpt4,llama2,chatglm,deepseekv3}. Multimodal extensions have carried these capabilities from text to images and video, and recent systems such as GPT-5 and Qwen3.5-Omni now handle several modalities natively inside a single model~\cite{blip2,llava,mrope,qwen25vl,llamavid,videollava,gpt5,qwen35omni}. Extending the same autoregressive paradigm to 3D content, however, remains far less mature. We argue that a 3D LLM must satisfy two requirements at once: 3D geometry should enter the model in an explicit form that preserves its spatial structure, and acquiring this new modality should not erode the general language ability of the underlying model.

Existing 3D LLMs fall short on one or both requirements. The first problem lies in how 3D geometry enters the model. ShapeLLM-Omni~\cite{shapellmomni} compresses each shape into VQVAE codebook indices, so the LLM observes only latent codes from which fine geometric detail has already been lost and whose spatial organization is no longer explicit, while LLaMA-Mesh~\cite{Wang2024a} serializes meshes as OBJ text, treating vertex coordinates as ordinary character strings and hiding the locality and hierarchy of the shape; related interfaces built on point-cloud encoders or mesh-derived tokens inherit similar trade-offs~\cite{threeDLLM,pointllm,fang2025meshllm}. The second problem lies in how the backbone absorbs the new modality. These systems typically adapt the LLM through full-model fine-tuning, which overwrites the pretrained text and vision-language behaviors: as we show in \cref{sec:experiments}, both ShapeLLM-Omni and LLaMA-Mesh lose a substantial fraction of their general language ability after 3D training. Existing methods therefore trade away either the spatial structure of the 3D input or the pretrained ability of the language model.

In this paper, we propose OctLLM, a 3D-native multimodal LLM built around two insights. First, an octree can serve as an explicit spatial language for 3D geometry, preserving locality and hierarchy hidden by latent codes or raw mesh text.
To handle sequence growth with depth, OctLLM exploits redundant fine-level occupancy by randomly emptying a fraction of second-to-last-level nodes and omitting their finest-level descendants, producing a \emph{Sparse Octree} (S-Octree). Each occupancy token remains anchored to its 3D coordinates and depth, so the LLM can directly model spatial structure when generating and interpreting shapes. With this representation, OctLLM better preserves part structure in generation and produces more accurate 3D descriptions than prior unified multimodal LLMs, including ShapeLLM-Omni (\cref{tab:3d_generation,tab:3d_understanding}). Sparsification also reduces training memory and time without sacrificing generation quality (\cref{app:ablation}).

Second, introducing 3D demands new capacity without overwriting language ability. Full-parameter fine-tuning is expressive but expensive and exposes every pretrained weight to 3D gradients. LoRA is cheaper but confines 3D learning to a low-rank update that, when merged, still changes the language pathway~\cite{lora}. OctLLM instead routes mesh and position-query tokens through full-rank 3D branches at a subset of blocks, training far fewer parameters than full fine-tuning, while text and image tokens stay on the frozen vision-language pathway. Shared self-attention connects the streams; separate mesh embeddings and output head keep occupancy prediction from competing with the language vocabulary. OctLLM reproduces the backbone exactly on MMLU and HellaSwag and stays within one point on GSM8K, whereas ShapeLLM-Omni still loses $34.5$ and $40.7$ points on the first two despite additional text training data (\cref{tab:language_ability}). Our contributions are as follows:

\begin{itemize}[leftmargin=16pt,itemsep=2pt]
    \item[-] We introduce OctLLM, which reads and writes S-Octrees as an explicit 3D language, using position-aware masks for generation and direct S-Octree input for understanding.
    \item[-] We develop a token-routed dual-stream transformer with full-rank 3D branches, mesh embeddings, and a mesh head beside a frozen text-image pathway.
    \item[-] OctLLM achieves state-of-the-art 3D generation and understanding among unified multimodal LLMs while preserving the backbone's language ability.
\end{itemize}

\section{Related Work} \label{sec:related}

\paragraph{Unified Multimodal Generation and Understanding.}
Casting generation and understanding as next-token prediction over one discrete sequence works well in 2D~\cite{Oord2017NeuralDR,Esser2021,Sun2024,Tian2024,Li2024}, where unified systems read and write several modalities with one backbone~\cite{chameleon2024,Wang2024,gpt4o,showo2025,transfusion2025,worldmodel2025}. Moving to 3D is harder because geometry must first become tokens~\cite{llm3dsurvey}, and existing 3D-native systems convert it in ways that discard structure: VQVAE codebook indices~\cite{Yin2023,Chen2024b,shapellmomni}, continuous point features~\cite{gpt4point}, or meshes serialized as coordinate text~\cite{Wang2024a,fang2025meshllm}, with related efforts adding part-level interfaces, reward-driven refinement, and expert-decoupled designs~\cite{partxmllm,rl3dr12025,uni3dar,unimesh2026,eva012026}. None of them lets the LLM operate on an explicit hierarchical 3D structure, and most adapt the backbone by full fine-tuning; OctLLM addresses both limitations.

\paragraph{Autoregressive 3D Generation and Octree Representations.}
Autoregressive shape generators differ mainly in how they serialize geometry. One line models the mesh itself, from PolyGen~\cite{Nash2020} and MeshGPT~\cite{Siddiqui2024} onward~\cite{Chen2024,Chen2024a,meshxl2024,edgerunner2025,Hao2024,deepmesh2025,pivotmesh2025,vertexregen2025,Weng2024}: surface fidelity is high, but sequences are long and rely on lossy VQ codebooks. A second line discretizes space hierarchically with octrees~\cite{Meagher1982}, which underlie efficient high-resolution 3D learning~\cite{Riegler2017,Wang2017,Wang2023,Xiong2024,Ren2024} and generators such as OctGPT~\cite{Wei2025} and Uni-3DAR~\cite{uni3dar}, together with adaptive and cross-scale tokenizations~\cite{oat2025,Zhang2024a,Zhang2024b,Qian2024}, where coarse structure and fine detail share one short coarse-to-fine sequence. Latent-diffusion, score-distillation, and feed-forward systems~\cite{Zhang2024,trellis,Zeng2022,Zhang2023a,Liu2023c,Nichol2022,Jun2023,Poole2022,Lin2023,Wang2023a,Hong2023,Tang2024} produce high-quality assets, but their continuous latents cannot serve as a language that an LLM reads and writes. OctLLM extends this line to a general-purpose LLM, keeping each token's coordinate and depth explicit.

\paragraph{3D Understanding and Language-Preserving Adaptation.}
Work on 3D understanding progressed from joint shape-text embeddings~\cite{y2seq2seq} to transferring 2D vision-language priors~\cite{Radford2021} onto point clouds~\cite{pointclip,pointclipv2,ulip} with stronger point encoders~\cite{Yu2022,Pang2022,Guo2021,Zhao2021}, and 3D MLLMs now connect such encoders, or raw point tokens, to an LLM for dialogue, grounding, and part-aware reasoning~\cite{threeDLLM,pointllm,shapellm,Yin2023,Wang2024a,fang2025meshllm,shapellmomni,partxmllm,sage3d2026}. How the backbone is adapted has received considerably less attention than what it perceives. Existing 3D LLMs acquire the new modality by updating the pretrained weights themselves, either by full fine-tuning or, where its cost is prohibitive, by low-rank adaptation~\cite{lora}. Either way the weights that language tokens depend on are modified, and the pretrained text and vision-language behavior degrades, as \cref{sec:experiments} confirms for existing 3D LLMs. In 2D, modality-decoupled architectures~\cite{mot2025,xfusion2025} instead route each modality through its own weights while sharing global self-attention; OctLLM brings this principle to 3D with full-rank branches that own their embeddings and output head, so 3D capability is acquired without degrading language ability.
\section{Method} \label{sec:method}

\begin{figure}[t]
  \centering
  \includegraphics[width=\linewidth]{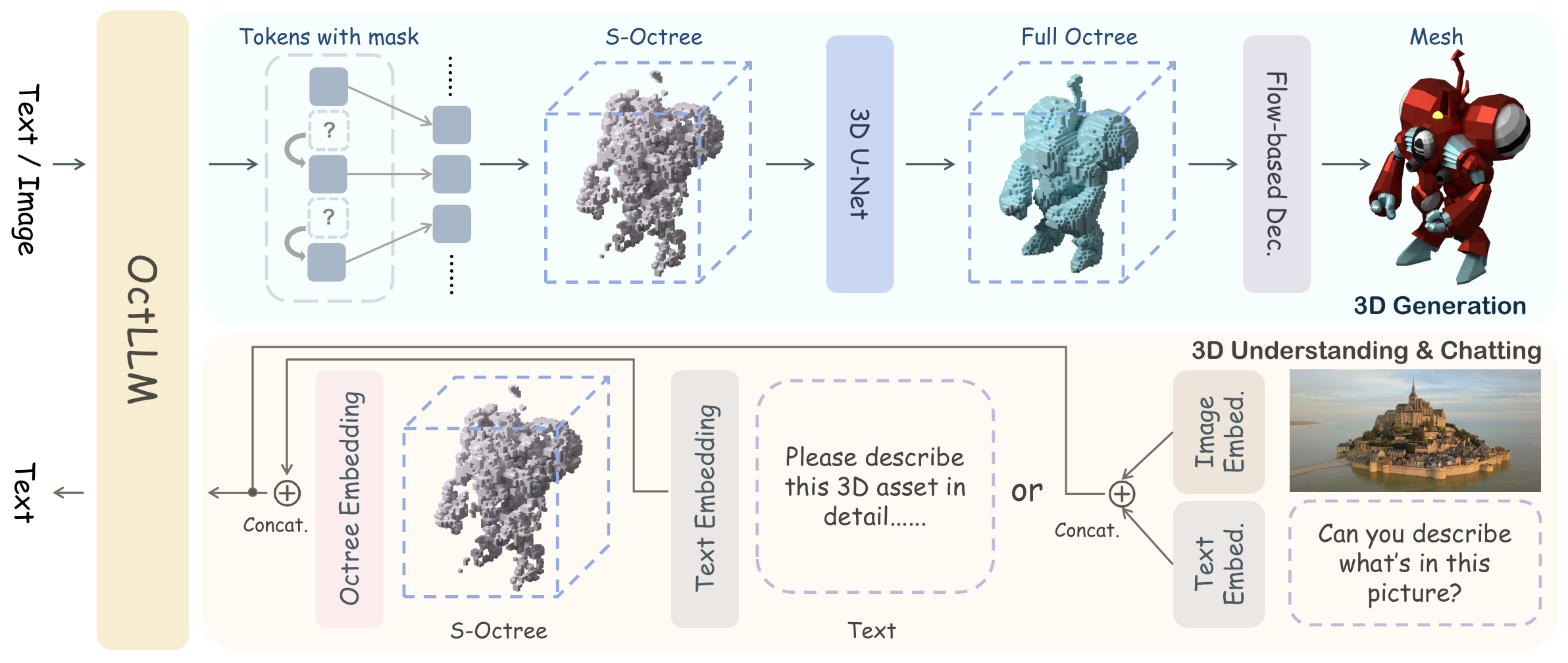}
  \caption{Overview of OctLLM. For generation, OctLLM autoregressively predicts a position-aware S-Octree sequence from a text or image condition, which a 3D U-Net completes into dense occupancy before a flow-based decoder, conditioned on the occupied voxels, produces the mesh. For 3D understanding, the same S-Octree is read with a text prompt as the 3D input language.}
  \label{fig:method_overview}
\end{figure}

\subsection{Octree-Based 3D Tokenization} \label{sec:tokenization}

OctLLM's pipeline, summarized in \cref{fig:method_overview}, begins by converting a mesh into language-model tokens using an octree.
Following OctGPT~\cite{Wei2025}, we encode the input mesh as an octree whose finest level matches the target voxel resolution and serialize the occupied 3D space with a Z-order traversal, so nearby 3D nodes tend to remain nearby in the one-dimensional sequence.
We then pack every $8$ consecutive binary occupancy bits into one byte-level mesh token whose value lies in $[0,255]$.
This reduces the sequence length by a factor of $8$ compared with a raw binary sequence, while keeping the representation discrete and faithful to the voxelized octree structure.
It also avoids the reconstruction artifacts that can appear when geometry is compressed into learned latent codes.

Although octrees already avoid subdividing empty space, serializing all occupied branches to a fine depth still yields a long sequence, making deeper octrees costly in training time and memory.
We further observe that the fine levels are highly redundant: many neighboring nodes have similar occupancy patterns, so randomly omitting a subset leaves the overall 3D structure nearly intact.
OctLLM exploits this redundancy by randomly setting a fixed fraction of nodes at the second-to-last level to empty, which also omits their finest-level descendants.
We call this additionally thinned representation a \emph{Sparse Octree} (S-Octree), distinguishing it from the octree's inherent empty-space sparsity (See \cref{fig:soctree}).
The removed fraction trades sequence length against the occupancy to be recovered.
A completion network later restores the omitted detail, as shown in \cref{fig:method_overview}, so the LLM focuses on global 3D layout and nontrivial shape parts while a geometry network handles dense local completion.
Because the thinned levels grow fastest with depth, the S-Octree shortens the sequence substantially while keeping the coarse structure that determines the object's identity, reducing memory and training time without sacrificing generation quality (\cref{app:ablation}).

\subsection{3D-Aware Dual-Stream Language Model} \label{sec:llm}

\begin{wrapfigure}[27]{r}{0.33\linewidth}
  \vspace{-1.5em}
  \centering
  \includegraphics[width=\linewidth]{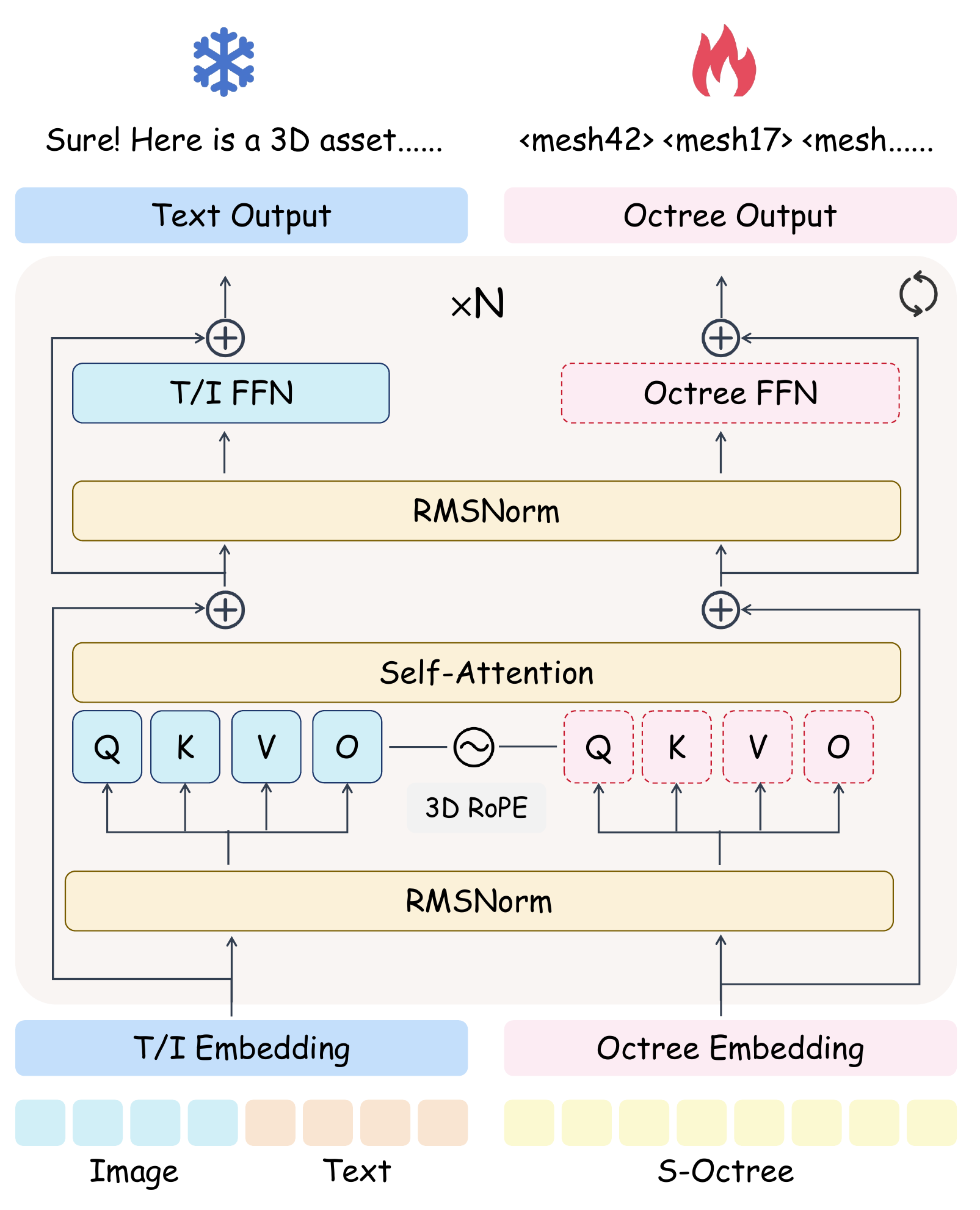}
  \caption{Token-routed dual-stream block. Text and image tokens retain the frozen modules, while S-Octree bytes and predictive \texttt{<MASK>} tokens use trainable 3D embeddings, projections, feed-forward branches, and a dedicated mesh head. Both streams interact through shared causal self-attention, so language, images, and geometry can exchange information.}
  \label{fig:dual_stream_architecture}
  \vspace{-0.8em}
\end{wrapfigure}

Adapting a pretrained LLM to 3D creates a capacity--preservation trade-off.
Full-parameter fine-tuning is expressive but costly and exposes pretrained transformations to 3D gradients.
LoRA is cheaper but restricts 3D learning to low-rank updates that, when merged, still alter the language pathway~\cite{lora}.
OctLLM instead routes octree bytes and predictive masks, with their 3D positions and depths, through full-rank trainable branches while text and image tokens remain on the unchanged pathway.
Mesh tokens receive independent input and output spaces, and shared self-attention connects the two streams, as shown in \cref{fig:dual_stream_architecture}.

\begin{figure}[t]
  \centering
  \begin{subfigure}[t]{0.66\linewidth}
    \centering
    \includegraphics[width=\linewidth]{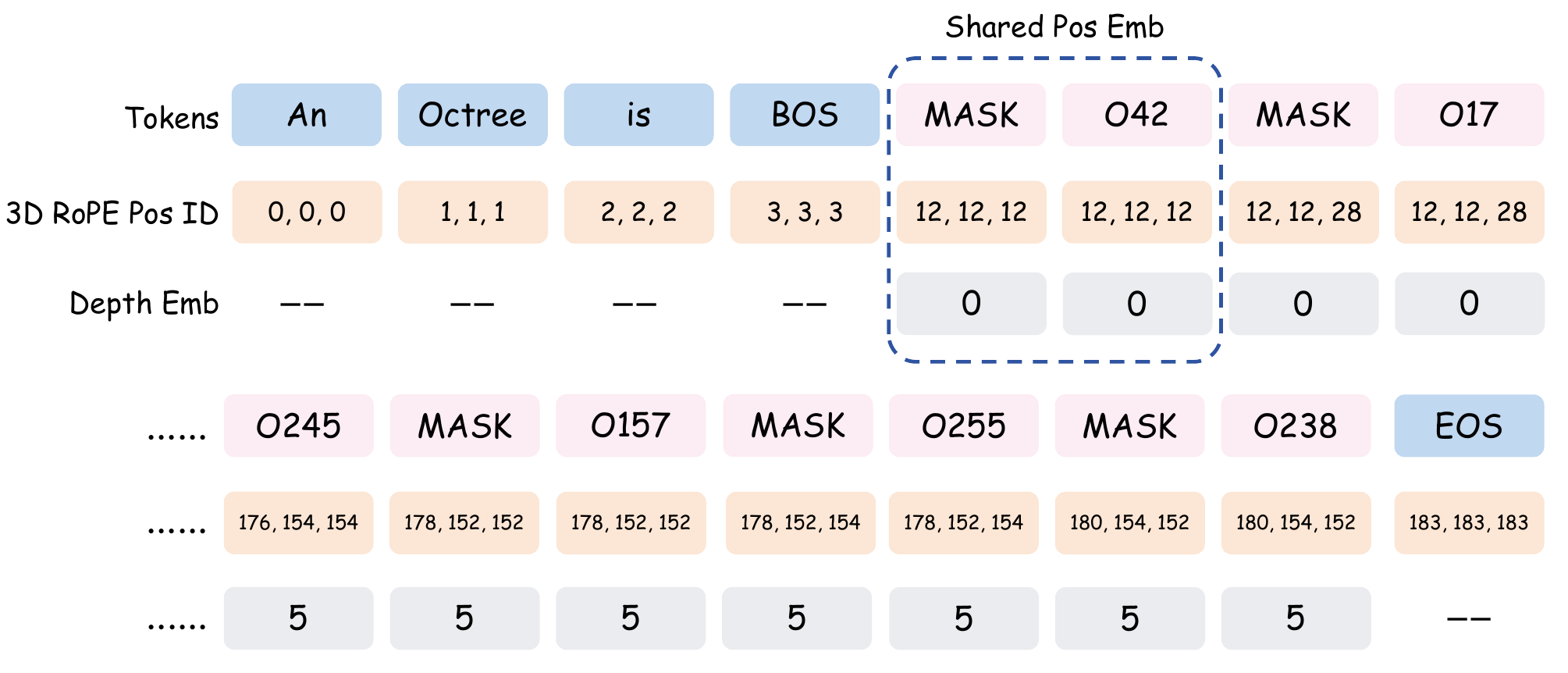}
    \caption{3D position and depth assignment.}
  \end{subfigure}
  \hfill
  \begin{subfigure}[t]{0.28\linewidth}
    \centering
    \includegraphics[width=\linewidth]{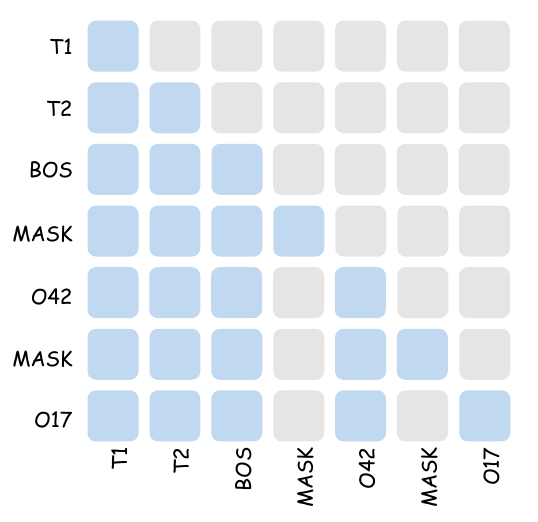}
    \caption{Attention mask.}
  \end{subfigure}
  \caption{Position-aware mask modeling. Each inserted \texttt{<MASK>} token shares the 3D rotary coordinates and depth embedding of the mesh-byte token that follows it, so the hidden state at the mask queries a known 3D location before predicting occupancy. Historical masks are suppressed as attention keys; text, boundary, and previous mesh-byte tokens stay visible.}
  \label{fig:position_mask_design}
\end{figure}

\paragraph{3D RoPE and Octree Depth Embedding.}
To let the language model see the spatial layout encoded in the octree sequence, we give every token a three-channel position identifier and apply an independent rotary encoding along each channel, so that a position is a point in a three-dimensional index space rather than a scalar offset.
Text and image tokens keep the sequential and raster assignments used by multimodal rotary embeddings~\cite{mrope}, but an octree token is not placed by its rank in the sequence: its three channels carry the 3D center coordinates of the node it encodes, rescaled per octree level as in OctGPT~\cite{Wei2025} so that all depths share one coordinate frame.
The model therefore sees both the serialized history and the spatial location of each occupancy token, which matters because tokens far apart in the sequence may still be adjacent in 3D space.
Coordinates alone, however, do not fully describe an octree token: a shallow node covers a large coarse region while a deep node covers a local surface detail.
Following OctGPT~\cite{Wei2025}, we therefore add a learnable depth embedding that tells the model the scale at which a token should be interpreted, so the same spatial neighborhood can be read differently at coarse and fine levels.
The left panel of \cref{fig:position_mask_design} illustrates how text positions, octree-node coordinates, mask tokens, and mesh-byte tokens are aligned in the expanded sequence.

\paragraph{Predictive Mask Tokens and Mask Suppression.}
During autoregressive octree generation, the next mesh byte has two meanings: where it is in 3D space and what occupancy pattern it contains, and asking the model to infer both at once makes the sequence harder to learn.
Inspired by Uni-3DAR~\cite{uni3dar}, OctLLM separates these two questions with a predictive mask token: before predicting a mesh byte, the model receives a \texttt{<MASK>} token carrying the target 3D position and octree depth, whose hidden state acts as a query asking what occupancy should appear at that location given the condition and the geometry generated so far.
Because these tokens are position queries rather than geometry observations, historical mask tokens are suppressed in attention while text, image, and previous mesh-byte tokens remain visible under the causal mask, as shown in the right panel of \cref{fig:position_mask_design}.
This keeps the model's context focused on the actual conditioning information and the generated geometry.
Details of training and inference with \texttt{<MASK>} token is given in \cref{app:training_procedure,app:inference_procedure}.

\paragraph{Token-Routed 3D Transformer Branches.}
The distribution of mesh bytes is tied to local geometry and has substantially higher entropy than that of natural-language tokens.
OctLLM therefore gives mesh-related tokens their own transformer transformations while keeping the pretrained pathway for text and image tokens, as shown in \cref{fig:dual_stream_architecture}.
The two streams still share causal self-attention, so language, images, and geometry can exchange information; only the token-local transformations are routed by token type.
Let $r_i \in \{0,1\}$ denote the route indicator for token $i$, where $r_i=1$ for mesh byte tokens and inserted \texttt{<MASK>} tokens, and $r_i=0$ for text, image/video, \texttt{<mesh\_bos>}, and \texttt{<mesh\_eos>} tokens.
For a routed block, the feed-forward output is computed by token-wise branch selection:
\begin{equation}
  \mathrm{FFN}_{\ell}^{\mathrm{route}}(h_i)
  =
  (1-r_i)\,\mathrm{FFN}_{\ell}^{\mathrm{base}}(h_i)
  +
  r_i\,\mathrm{FFN}_{\ell}^{\mathrm{3D}}(h_i),
  \label{eq:routed_ffn}
\end{equation}
where the base FFN is frozen and the 3D FFN is trainable.

The same idea applies to all four attention projections, with $Q$, $K$, and $V$ routed on the block input $h_i$ and the output projection routed on the attention result $a_i$ at the same token:
\begin{align}
  Q_i &= (1-r_i)\,W_Q h_i + r_i\,W_Q^{\mathrm{3D}} h_i, &
  K_i &= (1-r_i)\,W_K h_i + r_i\,W_K^{\mathrm{3D}} h_i, \notag \\
  V_i &= (1-r_i)\,W_V h_i + r_i\,W_V^{\mathrm{3D}} h_i, &
  O_i &= (1-r_i)\,W_O a_i + r_i\,W_O^{\mathrm{3D}} a_i,
  \label{eq:routed_qkvo}
\end{align}
where the pretrained projections $W_Q, W_K, W_V, W_O$ are frozen and their 3D counterparts are trainable.
The attention operation itself remains a single causal self-attention over the mixed sequence, so only the token-local projection and FFN transformations are separated.
The number of routed blocks trades 3D capacity against newly trained parameters, whereas their placement is not arbitrary: the placement study in \cref{app:ablation} shows that the 3D stream needs to reach early, middle, and late representations, and that deeper blocks contribute more per branch.
We therefore spread the routed blocks uniformly over the backbone and, as the budget grows, spend the additional branches in its second half.

\paragraph{Independent New-Token Embeddings and Output Head.}
OctLLM adds a small mesh vocabulary for octree bytes, mesh boundaries, and position-query masks.
These tokens have a different role from natural-language words, so they receive independent input embeddings and an independent output head while the original vocabulary keeps its pretrained embedding and language head, so 3D occupancy prediction does not compete with ordinary word prediction.
A remaining question is how the model should initiate mesh generation, since the pretrained language pathway has no prior ability to emit the newly introduced \texttt{<mesh\_bos>} token.
During instruction tuning we therefore format 3D-generation responses with short natural-language preambles (See the system prompt in \cref{app:inference_procedure}) that consistently lead into \texttt{<mesh\_bos>}, so the pretrained stream produces a familiar response pattern while the new head learns the high-probability transition from that pattern into the mesh sequence.

\subsection{Sparse Voxel Completion Network} \label{sec:unet}

The S-Octree predicted by OctLLM is compact but not yet a complete dense shape, so we decode it into a sparse occupancy grid at the finest octree resolution and use a 3D U-Net to infer the missing fine occupancy from the coarse, partial structure the LLM produced.
The encoder downsamples the grid by a factor of eight through four 3D convolutional layers of width $64$, $128$, $256$, and $256$, and the decoder mirrors it with four upsampling layers; the network is trained with binary cross-entropy over voxel occupancy.
The completed grid then supplies the occupied coordinates that a pretrained flow-based generator takes as its structural condition, together with the original text or image as its semantic condition, to produce the final textured mesh.

\subsection{Training Objectives} \label{sec:training}

OctLLM is trained on both 3D generation and 3D understanding data: for generation the model predicts mesh bytes from position-aware mask queries, and for understanding the S-Octree is part of the input and the supervision is the natural-language response.
Only content tokens are prediction targets, since mask tokens provide where to predict while mesh bytes or words provide what to predict.
Mask tokens are consequently inserted only in generation samples; in understanding samples the S-Octree is already given, so no location needs to be queried and the mesh sequence is fed without them, which also avoids doubling the length of the 3D input.

The loss reflects the same separation between words and mesh bytes: text targets use the language vocabulary distribution and mesh-byte targets use the mesh vocabulary distribution, so occupancy prediction is not diluted by irrelevant word logits.
Let $\mathcal{T}_{\mathrm{text}}$ denote supervised text-token positions and $\mathcal{T}_{\mathrm{mesh}}$ denote supervised mesh-byte positions.
The training objective is
\begin{equation}
  \mathcal{L}_{\text{LLM}} =
  -\frac{1}{|\mathcal{T}_{\mathrm{text}}| + |\mathcal{T}_{\mathrm{mesh}}|}
  \left(
    \sum_{t \in \mathcal{T}_{\mathrm{text}}}
    \log P_{\mathrm{vocab}}\!\left(x_t \mid x_{<t}\right)
    +
    \sum_{t \in \mathcal{T}_{\mathrm{mesh}}}
    \log P_{\mathrm{mesh}}\!\left(x_t \mid x_{<t}\right)
  \right),
  \label{eq:lm_loss}
\end{equation}
where $P_{\mathrm{vocab}}$ is the language vocabulary distribution and $P_{\mathrm{mesh}}$ is the mesh-byte distribution.
Both generation and understanding data are mixed during training, so the 3D branches learn from mesh-byte prediction while the shared attention pathway also supports 3D-conditioned language responses.

\section{Experiments} \label{sec:experiments}

\subsection{Implementation Details}

\paragraph{Dataset.}
We train OctLLM on the union of the Sketchfab subset of Objaverse-XL~\cite{objaversexl}, HSSD~\cite{hssd}, ABO~\cite{abo}, and ShapeNet~\cite{Chang2015}, processed following TRELLIS~\cite{trellis}: each asset is rendered from 24 Blender views, one of which becomes its image condition, and four fixed views are captioned by Qwen3.5-9B~\cite{qwen35}.
Each caption serves as both the text-to-3D prompt and the 3D-to-text reference, so both directions of the 3D language are supervised against the same description.
Excluding assets in the PointLLM-200 test set, this gives 195K assets and 585K instructions; the completion network is trained separately on 175K sparse--complete occupancy pairs from the same Objaverse-XL subset.

\paragraph{Training Details.}
The S-Octree uses six octree levels, so its finest grid is $64^3$, and is sparsified by dropping second-to-last-level nodes with probability $0.5$; TRELLIS~\cite{trellis} serves off the shelf as the flow-based decoder of \cref{sec:unet}.
The text-image pathway is Qwen2.5-VL-7B-Instruct~\cite{qwen25vl}, whose multimodal rotary embedding~\cite{mrope} already gives every token three position channels, which our 3D coordinates reuse.
We route 3D branches through 12 of its 28 decoder blocks, at the 0-indexed layers $\{0,4,8,12,14,16,18,20,22,24,26,27\}$; this adds about 2.80B trainable parameters, roughly a quarter of the total and about a third of what full fine-tuning updates. \cref{app:ablation} motivates this layout and the 8-branch ablation variant.
Each branch is initialized from the block it accompanies, which stabilizes early training.
We freeze the language backbone and the vision encoder and train only the routed 3D branches, the mesh embeddings and mesh head, and the octree depth encoding; the mesh vocabulary adds 256 occupancy-byte tokens plus \texttt{<mesh\_bos>}, \texttt{<mesh\_eos>}, and \texttt{<MASK>} in their own embedding space.
Training runs for 69K steps on 16 H20 GPUs and the completion network for 155K steps on 8 RTX 5090 GPUs, both with AdamW at an effective batch size of 64; \cref{app:protocol} lists the remaining optimization settings.

\subsection{Quantitative Comparison}

\paragraph{Evaluation Protocol.}
Generation is scored on 1K held-out Toys4K assets~\cite{stojanov2021toys4k} by render-based FID and KID with Inception-V3~\cite{szegedy2016inception} and DINOv2~\cite{oquab2024dinov2} features plus CLIP~\cite{Radford2021} condition alignment, with untextured baseline outputs textured by the same tool.
Understanding is evaluated on the PointLLM-200 captioning benchmark~\cite{pointllm}.
Following PointLLM, we report Sentence-BERT and SimCSE similarity plus two Qwen3.8-Max judges~\cite{qwen38max}: GPT-ref compares a caption against the human reference, and GPT-img withholds it, scoring the caption against four renders of the object.
$n$-gram metrics are reported only in \cref{app:protocol}, since PointLLM shows them to be unreliable for 3D captioning.
All baselines are evaluated with their publicly released model weights and code, and every test asset is converted into the 3D input format that the corresponding method expects.
Throughout we keep two families of baselines apart: \emph{task-specific} models trained for one task with one model per modality, and \emph{multimodal LLMs} carrying generation, understanding, and language in one backbone.

\begin{figure}[t]
  \centering
  \includegraphics[width=\linewidth,trim=6 22 0 14,clip]{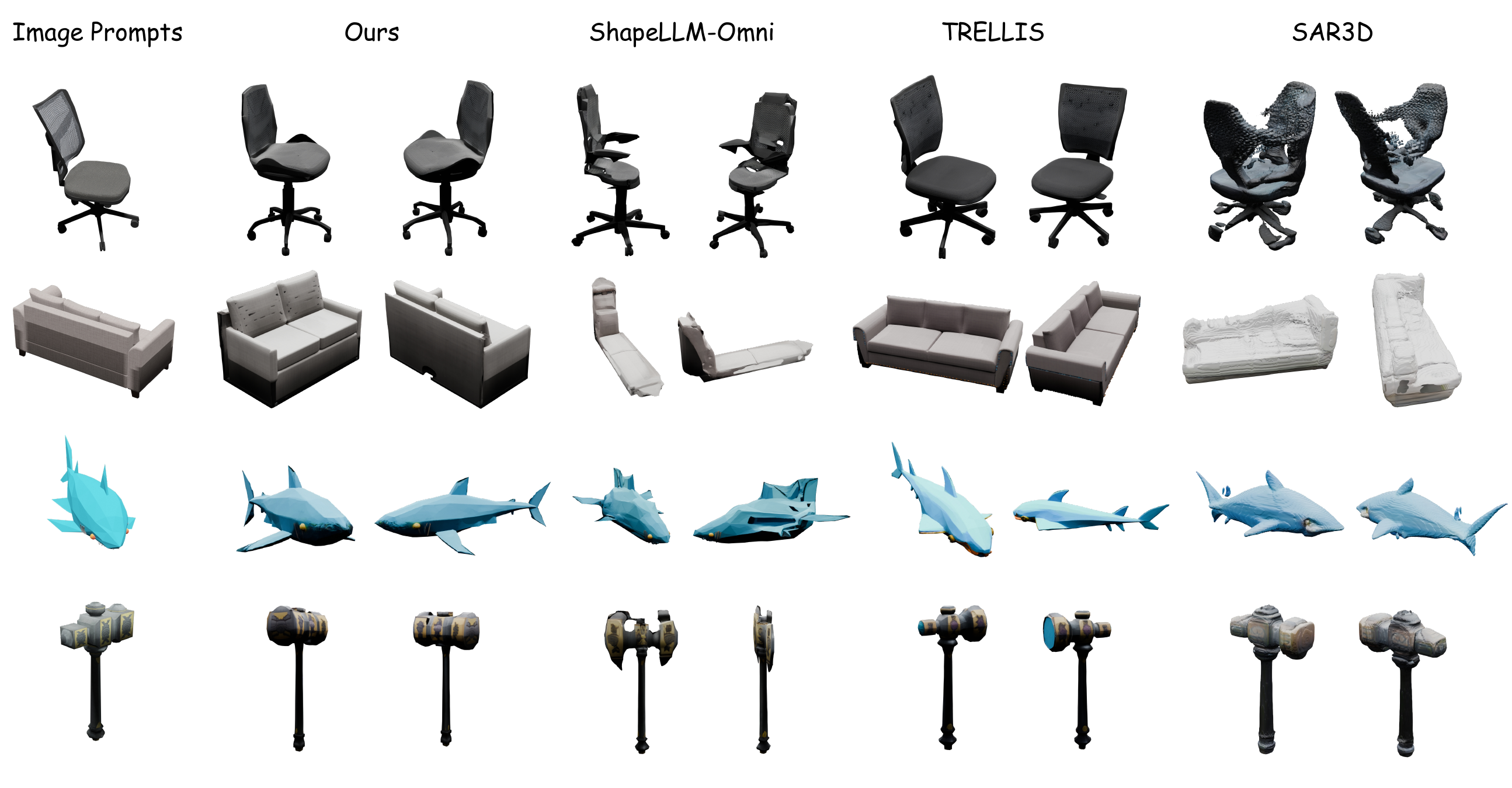}
  \caption{\textbf{Image-to-3D generation.} Each asset is shown from two viewpoints. OctLLM follows the structure and part layout of the conditioning image, while ShapeLLM-Omni and SAR3D produce geometry of lower quality.}
  \label{fig:qualitative_image}
\end{figure}

\paragraph{3D Generation Quality.}
As reported in \cref{tab:3d_generation}, OctLLM outperforms all multimodal LLMs on image-conditioned generation, reducing Inception FID by $17.4\%$ and KID by $45\%$ relative to ShapeLLM-Omni.
Under text conditioning, OctLLM leads the same family on Inception FID and both KID scores, where it also surpasses the task-specific octree generator OctGPT, and performs comparably to ShapeLLM-Omni on DINOv2 FID and CLIP.
TRELLIS attains the best scores overall, and OctLLM ranks second on most of the metrics.
This gap is expected: TRELLIS is a dedicated 3D generator, whereas OctLLM produces geometry from a frozen-backbone LLM that also describes shapes and retains its language ability. Notably, ShapeLLM-Omni and AR3D-R1 likewise use TRELLIS as the flow-based mesh decoder, so the gain of OctLLM over them comes from our new design of 3D representation and pipeline, which produces higher-quality voxels for TRELLIS.

\begin{table}[t]
  \vspace{-0.4em}
  \caption{\textbf{Image-to-3D and text-to-3D generation on Toys4K.} FID/KID are computed on rendered views, with KID scaled by $\times 100$, and CLIP measures condition alignment. Each modality block lists only the methods supporting that conditioning, split into task-specific 3D generators and multimodal LLMs; within a block, the \colorbox{bestcolor}{best} and \colorbox{secondcolor}{second-best} result per column are highlighted.}
  \label{tab:3d_generation}
  \vspace{-0.5em}
  \tablestyle{5pt}{1.0}
  \resizebox{\linewidth}{!}{
  \begin{tabular}{@{}l ccccc@{}}
    \toprule
    Method & FID (Inc.)~$\downarrow$ & KID (Inc.)~$\downarrow$ & FID (DINOv2)~$\downarrow$ & KID (DINOv2)~$\downarrow$ & CLIP~$\uparrow$ \\
    \midrule
    \rowcolor{graycolor}
    \multicolumn{6}{c}{\textit{Image-to-3D}} \\
    \midrule
    \multicolumn{6}{@{}l}{\textit{Task-specific 3D generators}} \\
    TRELLIS~\cite{trellis}       & \best{18.15} & \best{0.13} & \best{48.47} & \best{2.06} & \best{92.86} \\
    SAR3D~\cite{Chen2024b}       & 67.42 & 3.16 & 206.79 & 20.56 & 84.46 \\
    \cmidrule(lr){1-6}
    \multicolumn{6}{@{}l}{\textit{Multimodal LLMs}} \\
    ShapeLLM-Omni~\cite{shapellmomni} & 38.21 & 1.02 & 145.14 & 9.98 & 86.75 \\
    \textbf{OctLLM (Ours)}       & \second{31.58} & \second{0.56} & \second{135.58} & \second{7.94} & \second{87.31} \\
    \midrule
    \rowcolor{graycolor}
    \multicolumn{6}{c}{\textit{Text-to-3D}} \\
    \midrule
    \multicolumn{6}{@{}l}{\textit{Task-specific 3D generators}} \\
    TRELLIS~\cite{trellis}       & \best{30.90} & \best{0.34} & \best{113.16} & \best{3.54} & \best{24.71} \\
    SAR3D~\cite{Chen2024b}       & 86.87 & 4.51 & 296.74 & 29.26 & 18.70 \\
    OctGPT~\cite{Wei2025}        & 46.40 & 1.13 & 228.81 & 15.92 & 19.70 \\
    \cmidrule(lr){1-6}
    \multicolumn{6}{@{}l}{\textit{Multimodal LLMs}} \\
    ShapeLLM-Omni~\cite{shapellmomni} & 52.36 & 1.38 & \second{169.28} & 10.34 & \second{22.46} \\
    LLaMA-Mesh~\cite{Wang2024a}  & 66.68 & 2.78 & 275.94 & 23.62 & 17.90 \\
    AR3D-R1~\cite{rl3dr12025}    & 57.65 & 1.78 & 187.33 & 12.95 & 21.89 \\
    \textbf{OctLLM (Ours)}       & \second{45.61} & \second{0.98} & 175.58 & \second{9.01} & 22.30 \\
    \bottomrule
  \end{tabular}}
  \vspace{-0.3em}
\end{table}

\begin{table}[t]
  \vspace{-0.4em}
  \caption{\textbf{3D understanding and language ability.} (a) Captioning on the 200-object PointLLM benchmark, judged by Qwen3.8-Max; the rule separates task-specific understanding models from multimodal LLMs, and the \colorbox{bestcolor}{best} and \colorbox{secondcolor}{second-best} result per column are highlighted; \cref{tab:3d_understanding_full} adds PointLLM-13B and $n$-gram metrics. (b) General language benchmarks under greedy decoding, with the backbone in gray as a reference rather than a competitor and the \colorbox{bestcolor}{best} result per column highlighted. Higher is better throughout.}
  \label{tab:understanding_language}
  \vspace{-0.5em}
  \begin{subtable}[t]{0.49\linewidth}
    \caption{3D understanding on PointLLM-200.}
    \label{tab:3d_understanding}
    \tablestyle{4pt}{1.15}
    \resizebox{\linewidth}{!}{
    \begin{tabular}{@{}l cccc@{}}
      \toprule
      Method & S-BERT~$\uparrow$ & SimCSE~$\uparrow$ & GPT-ref~$\uparrow$ & GPT-img~$\uparrow$ \\
      \midrule
      PointLLM-7B            & \best{48.05} & \best{48.63} & \best{46.06} & \second{45.63} \\
      ShapeLLM-7B            & 37.27 & 37.10 & 15.92 & 30.23 \\
      \cmidrule(lr){1-5}
      LLaMA-Mesh             & 30.44 & 31.08 & 15.60 & 17.16 \\
      ShapeLLM-Omni          & 36.08 & 34.28 & 22.27 & 18.14 \\
      \textbf{OctLLM (Ours)} & \second{38.80} & \second{38.48} & \second{26.61} & \best{46.88} \\
      \bottomrule
    \end{tabular}}
  \end{subtable}
  \hfill
  \begin{subtable}[t]{0.49\linewidth}
    \caption{Language ability preservation.}
    \label{tab:language_ability}
    \tablestyle{4pt}{1.38}
    \resizebox{\linewidth}{!}{
    \begin{tabular}{@{}l cccc@{}}
      \toprule
      Method & MMLU~$\uparrow$ & HellaSwag~$\uparrow$ & GSM8K~$\uparrow$ & IFEval~$\uparrow$ \\
      \midrule
      \rowcolor{graycolor}
      Qwen2.5-VL-7B            & 64.65 & 67.44 & 83.54 & 75.41 \\
      \cmidrule(lr){1-5}
      ShapeLLM-Omni            & 30.17 & 26.74 & 27.45 & 58.04 \\
      LLaMA-Mesh               & 51.84 & 34.03 & 52.69 & 66.54 \\
      \textbf{OctLLM (Ours)}   & \best{64.65} & \best{67.44} & \best{83.01} & \best{69.87} \\
      \bottomrule
    \end{tabular}}
  \end{subtable}
  \vspace{-0.8em}
\end{table}

\paragraph{3D Understanding Quality.}
Among multimodal LLMs, OctLLM leads on every metric of \cref{tab:3d_understanding}, by more than 28 points under the render-grounded judge, since it reads the coordinate-anchored occupancy sequence which carries explicit 3D information.
Across all methods, OctLLM obtains the best GPT-img score and ranks second on the other three metrics, outperforming ShapeLLM-7B throughout and trailing only PointLLM-7B under GPT-ref and the embedding scores.
The remaining gap reflects what each input carries: PointLLM reads colored point clouds and can therefore report material and color, whereas the S-Octree encodes geometry alone, so OctLLM describes shape and part structure and says nothing about appearance.

\paragraph{Language Ability Preservation.}
OctLLM reproduces the backbone exactly on both multiple-choice benchmarks and differs by less than one point on GSM8K, whereas the baselines lose between a third and a half of its accuracy (\cref{tab:language_ability}).
This follows from the architecture rather than from data balancing: the pretrained parameters are frozen and text tokens never enter a 3D branch, so a text-only conversation reproduces the backbone's computation exactly.
The residual drop on IFEval is caused by the mesh head occasionally opening a 3D sequence in the middle of an answer, and is eliminated by masking the mesh vocabulary at inference time.

\subsection{Qualitative Comparison}

\begin{figure}[t]
  \centering
  \includegraphics[width=0.99\linewidth,trim=16 10 2 12,clip]{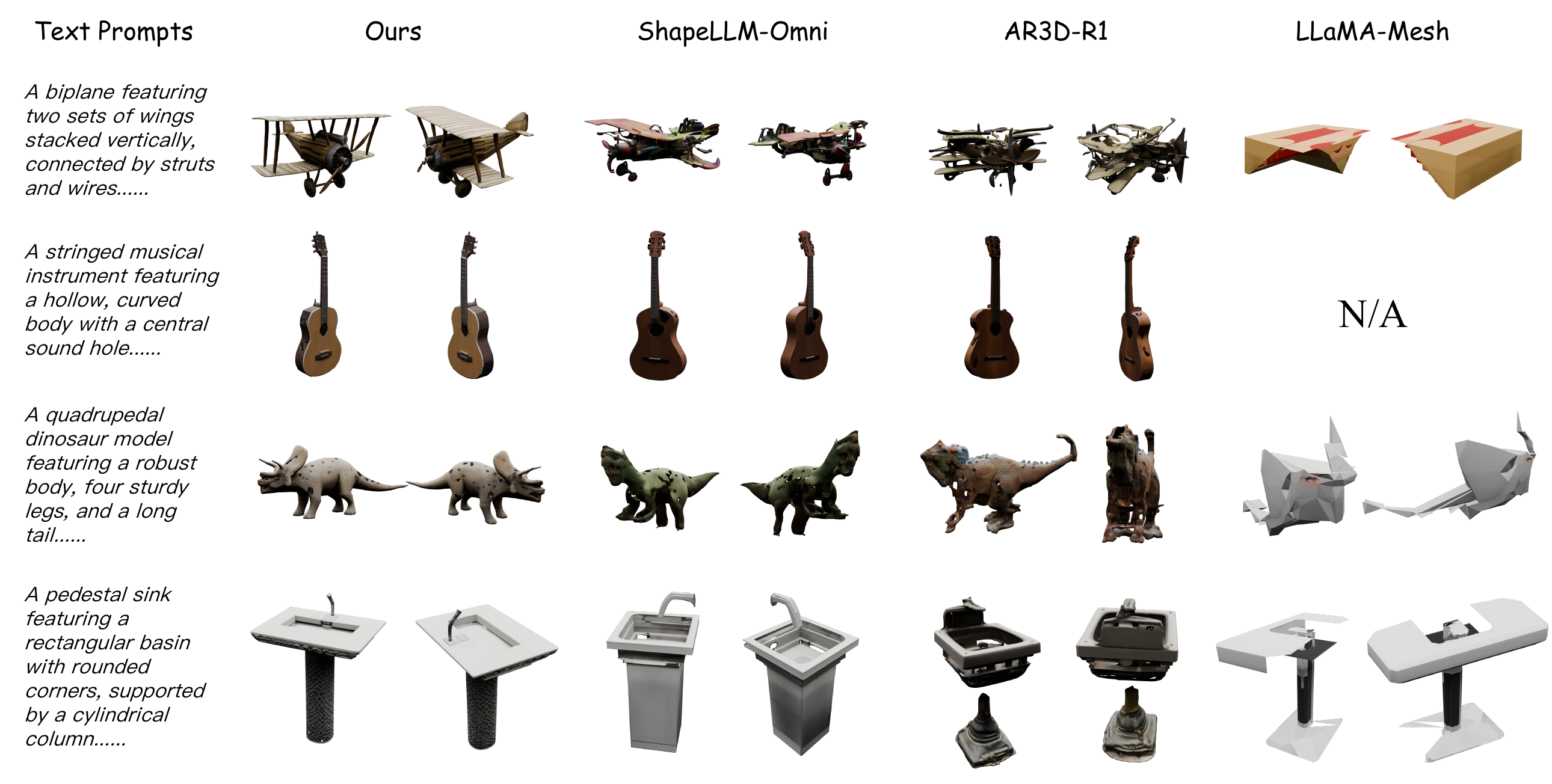}
  \vspace{-0.8em}
  \caption{\textbf{Text-to-3D generation.} Prompts are truncated for space and describe how parts are composed; \cref{app:complete_qualitative_text} provides the complete conditions. ``N/A'' marks a prompt for which LLaMA-Mesh returned no valid mesh.}
  \label{fig:qualitative_text}
  \vspace{-1.4em}
\end{figure}

\paragraph{Image-to-3D Generation.}
\Cref{fig:qualitative_image} compares OctLLM with ShapeLLM-Omni, TRELLIS, and SAR3D on four held-out assets, each rendered from two viewpoints.
SAR3D recovers only a coarse silhouette, and its noisy surfaces lose the finer parts of each object.
ShapeLLM-Omni produces cleaner surfaces but still does not preserve part-level structure.
OctLLM reproduces both the global layout and the individual parts of the conditioning image, and is visually closest to TRELLIS, which also serves as the structure-conditioned decoder of our pipeline.

\paragraph{Text-to-3D Generation.}
\Cref{fig:qualitative_text} compares OctLLM with ShapeLLM-Omni, AR3D-R1, and LLaMA-Mesh on prompts that name parts and state how they are composed.
LLaMA-Mesh returns no valid mesh for the stringed instrument, and ShapeLLM-Omni and AR3D-R1 recover a recognizable object category but follow the part composition stated in the prompt less closely than OctLLM.
OctLLM matches these constraints most closely, producing two stacked wings joined by struts, four legs with a long tail, and a rectangular basin on a cylindrical column.
With geometry as an explicit 3D sequence, OctLLM controls finer structure and better matches the text prompt.

\paragraph{3D Understanding.}
\Cref{fig:qualitative_understanding} in the appendix shows that OctLLM describes 3D inputs in more detail and with fewer errors than the compared methods.

\FloatBarrier

\section{Conclusion} \label{sec:conclusion}

We presented OctLLM, a multimodal LLM that treats a sparse octree as an explicit 3D language: geometry enters and leaves the model as occupancy bytes anchored to 3D coordinates and octree depths, routed through trainable branches placed beside a frozen text-image pathway. Our experiments support both design choices. Keeping 3D structure explicit rather than compressed yields the best generation and 3D understanding among unified multimodal LLMs, while adding 3D capacity in separate parameters reproduces the backbone exactly on general language benchmarks, where 3D LLMs that update the pretrained weights lose a third to a half of their accuracy.

\subsection*{AI use statement}

In this work, we used GPT-5.6 Sol to assist with drafting and language editing,
to identify and summarize relevant literature for the related-work section, and
to implement portions of the codebase. We used locally deployed Qwen3.5-9B to
generate captions that form part of the training data, as described in
\cref{sec:experiments} and \cref{app:data_construction}, and Qwen3.8-Max through
Alibaba's official API as an automated evaluation judge (\cref{app:protocol}). We did not use generative AI to
formulate the core research idea or mathematical claims, design or provide
feedback on the methodology or experiments, or interpret the results and draw
conclusions; proof-related tasks are not applicable to this work. The authors
reviewed all AI-assisted text, literature references, code, and generated data,
and checked the code and data for correctness and suitability for the reported
experiments. We take full responsibility for all content of this work, including AI-assisted text, claims, code, and data.

\subsection*{Reproducibility statement}

The main paper and appendix provide the information needed to reproduce our
results. \Cref{sec:method} specifies the representation, architecture, and
training objectives, while \cref{sec:experiments} describes
the datasets, training setup, and evaluation protocol. \Cref{app:procedures} and
\cref{app:protocol} further detail the end-to-end training and inference procedures,
data preprocessing and caption annotation, optimization settings, rendering
protocol, and metric computation. We have publicly released the source code
and model weights. The release includes all
scripts needed to reproduce the reported experiments and results, together with
instructions for using the code and carrying out the complete reproduction
workflow. Both are accessible from our project page:
\url{https://plurato.github.io/OctLLM-page/}.

\subsubsection*{Acknowledgments}
We thank Nachuan Duan, Runnan Hou, and Mingyi Zhang for their suggestions,
discussions, and help with auxiliary experiments during this work.

\bibliography{src/ref/reference}
\bibliographystyle{iclr2027_conference}

\clearpage
\appendix
\raggedbottom
\setcounter{figure}{0}
\setcounter{table}{0}
\setcounter{algorithm}{0}
\setcounter{equation}{0}
\renewcommand{\thefigure}{S\arabic{figure}}
\renewcommand{\thetable}{S\arabic{table}}
\renewcommand{\thealgorithm}{S\arabic{algorithm}}
\renewcommand{\theequation}{S\arabic{equation}}
\makeatletter
\@ifundefined{theHfigure}
  {\newcommand{\theHfigure}{supp.figure.\arabic{figure}}}
  {\renewcommand{\theHfigure}{supp.figure.\arabic{figure}}}
\@ifundefined{theHtable}
  {\newcommand{\theHtable}{supp.table.\arabic{table}}}
  {\renewcommand{\theHtable}{supp.table.\arabic{table}}}
\@ifundefined{theHalgorithm}
  {\newcommand{\theHalgorithm}{supp.algorithm.\arabic{algorithm}}}
  {\renewcommand{\theHalgorithm}{supp.algorithm.\arabic{algorithm}}}
\@ifundefined{theHequation}
  {\newcommand{\theHequation}{supp.equation.\arabic{equation}}}
  {\renewcommand{\theHequation}{supp.equation.\arabic{equation}}}
\makeatother
\section*{APPENDIX -- Octrees as an Explicit 3D Language}

This supplement is organized as follows. \Cref{app:procedures} details the training and inference procedures, \cref{app:protocol} the experimental setup and evaluation protocol, and \cref{app:additional_generation} additional qualitative results, the judge prompts, and an inference-efficiency comparison. \Cref{app:ablation} isolates the contribution of each design choice, and \cref{app:limitations} discusses the remaining limitations and future directions.

\section{Training and Inference Procedures} \label{app:procedures}

This section makes the end-to-end procedures behind \cref{sec:training,sec:unet} explicit. We use $c$ for a text or image condition, $M$ for a mesh, $S$ for its serialized S-Octree, and $G$ and $G_S$ for the complete and sparse occupancy grids, respectively. The language model and the completion network are optimized independently: the former learns the shared 3D language used for generation and understanding, whereas the latter only learns to restore occupancy removed by S-Octree sparsification.

\subsection{Training Procedure} \label{app:training_procedure}

For a generation sample, we construct the target S-Octree by voxelizing the mesh, pruning nodes at the designated fine level, traversing the remaining octree in Z-order, and packing every eight occupancy bits into one mesh-byte token. This sparsification is performed once, when the instruction-tuning dataset is built: the random node emptying is drawn a single time per asset and the resulting S-Octree is stored, so each mesh has one fixed byte sequence throughout training and \cref{eq:lm_loss} is applied to a deterministic target. Teacher forcing inserts a \texttt{<MASK>} query immediately before every target mesh byte. The query is assigned the position and depth of that byte, and historical mask queries are suppressed as attention keys. Supervision is applied to the mesh-byte targets through the mesh head. For an understanding sample, the S-Octree is instead part of the input, without inserted mask queries, and only the natural-language answer is supervised through the language head. Both sample types follow the objective in \cref{eq:lm_loss}; \cref{alg:training} summarizes this procedure together with the separate training of the completion network on paired sparse and complete occupancy grids.

\begin{algorithm}[H]
  \caption{Training OctLLM and the sparse voxel completion network.}
  \label{alg:training}
  \begin{algorithmic}[1]
    \Require Pretrained text--image parameters $\theta_0$; trainable 3D parameters $\theta_{3D}$; completion network $U_{\phi}$; generation and understanding datasets $\mathcal{D}_{\mathrm{gen}}$ and $\mathcal{D}_{\mathrm{und}}$
    \Ensure Trained $\theta_{3D}$ and $\phi$
    \State Precompute $S \gets \operatorname{Serialize}(\operatorname{Sparsify}(\operatorname{Octree}(M)))$ for every mesh $M$ \Comment{once, before training}
    \State Freeze $\theta_0$
    \For{each mixed minibatch $B \subset \mathcal{D}_{\mathrm{gen}} \cup \mathcal{D}_{\mathrm{und}}$}
      \For{each sample in $B$ with its stored S-Octree $S$}
        \If{the sample is $(c,S) \in \mathcal{D}_{\mathrm{gen}}$}
          \State Insert a position-aware \texttt{<MASK>} before every byte in $S$
          \State Supervise the S-Octree bytes conditioned on $c$
        \Else
          \State Provide $S$ without mask queries and supervise the text answer
        \EndIf
      \EndFor
      \State Update $\theta_{3D}$ with $\mathcal{L}_{\mathrm{LLM}}$ in \cref{eq:lm_loss}; keep $\theta_0$ fixed
    \EndFor
    \For{each mesh $M$ in the completion dataset}
      \State $G \gets \operatorname{Voxelize}(\operatorname{Octree}(M))$
      \State $G_S \gets \operatorname{Voxelize}(\operatorname{Sparsify}(\operatorname{Octree}(M)))$
      \State Update $\phi$ with $\mathcal{L}_{\mathrm{comp}}=\operatorname{BCE}(U_{\phi}(G_S),G)$
    \EndFor
  \end{algorithmic}
\end{algorithm}

\subsection{Generation Procedure} \label{app:inference_procedure}

\paragraph{System prompt.}
OctLLM uses the following system prompt for both generation and understanding:

\begin{PromptBox}[samepage=true]
You are a helpful assistant specialized in 3D asset generation and
understanding, image understanding and chatting. When you are asked
to generate a 3D asset from an image, you should reply in formats
like: "I've produced a 3D model based on the image: ", "Based on the
image, here's the 3D mesh asset I've created: ", etc. And if you are
asked to generate a 3D asset from a text description, you should
reply in formats like: "I've produced a 3D mesh asset based on your
description: ", "Of course! I've generated a 3D mesh based on your
text prompt: ", etc. If you are asked to describe a 3D asset, just
describe it in detail. The text can be varied, but the colon ":"
must be included.
\end{PromptBox}

The prompt establishes a single assistant role spanning 3D generation, 3D understanding, image understanding, and ordinary dialogue. For generation, it also standardizes the natural-language preamble to end in a colon, matching the instruction-tuning format and creating a reliable transition to \texttt{<mesh\_bos>}, which starts constrained octree decoding. Without this task and output-format contract, the backbone may answer conversationally or terminate before emitting the mesh boundary, leaving no dependable handoff to the structured generation loop.

\begin{figure}[H]
  \centering
  \includegraphics[width=\linewidth]{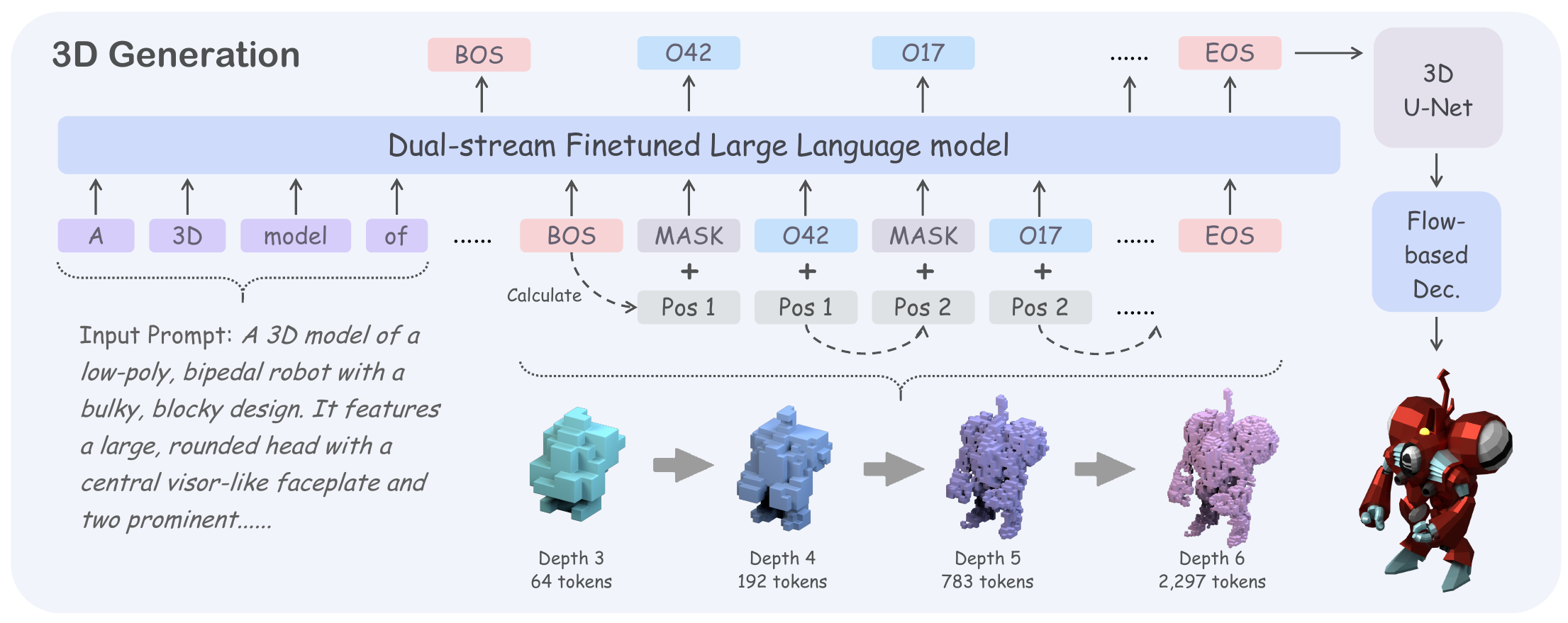}
  \caption{Text-conditioned generation at inference time. The prompt is answered with a short natural-language preamble that leads into \texttt{<mesh\_bos>}, after which each mesh byte is produced by first attaching the next node's 3D position to a transient \texttt{<MASK>} query and then sampling only the occupancy value at that position. The partial octree grows coarse to fine, and the completed grid is passed to the completion network and the flow-based decoder.}
  \label{fig:inference_pipeline}
\end{figure}

\Cref{fig:inference_pipeline} illustrates the loop described below. At inference time, the partially generated octree uniquely determines the next node in the coarse-to-fine Z-order traversal. OctLLM therefore does not need to generate a coordinate token: it attaches the next node's 3D position and depth to a transient \texttt{<MASK>} query and samples only its occupancy byte. Sampling is logit-masked to the 256 valid occupancy bytes, so the model cannot emit \texttt{<mesh\_eos>}, \texttt{<MASK>}, or a word token in the middle of the traversal. Once the byte is sampled, the key--value entries of the mask query are physically removed from the KV cache; this realizes the mask suppression of \cref{sec:llm} at inference time, and the cached context grows only by the sampled byte at each step. The sampled byte is unpacked into eight split decisions and immediately updates the partial octree, which in turn determines the following query. Generation stops when the target octree depth is complete, rather than relying on the language model to infer structural termination.

The completed byte sequence is decoded into the sparse occupancy grid $G_S$ and passed through the 3D U-Net. After binarizing its occupancy prediction at a threshold of $0.5$, the occupied voxels are converted to the sparse coordinates expected by TRELLIS. TRELLIS uses these coordinates as the structural condition and the original text or image as the semantic condition, samples its structured latent representation, and decodes the final mesh. \Cref{alg:inference} summarizes this complete path.

\begin{algorithm}[H]
  \caption{OctLLM generation from a text or image condition.}
  \label{alg:inference}
  \begin{algorithmic}[1]
    \Require Condition $c$; OctLLM $F_{\theta_0,\theta_{3D}}$; completion network $U_{\phi}$; target octree depth $D$; TRELLIS pipeline $T$
    \Ensure S-Octree sequence $S$ and decoded mesh $\widehat{M}$
    \State Generate the response prefix and \texttt{<mesh\_bos>} from $c$
    \State Initialize an empty S-Octree sequence $S$ and partial octree $O$
    \While{$O$ is not complete through depth $D$}
      \State $(\mathbf{p},d) \gets \Call{NextNode}{O}$
      \State $q \gets \texttt{<MASK>}(\mathbf{p},d)$
      \State $b \sim P_{\mathrm{mesh}}(\cdot \mid c,S,q)$ \Comment{logits masked to the 256 occupancy bytes}
      \State Remove the key--value entries of $q$ from the KV cache
      \State Append mesh byte $b$ to $S$ and update $O$ with $\Call{Unpack8}{b}$
    \EndWhile
    \State Append \texttt{<mesh\_eos>} and decode $S$ into sparse occupancy $G_S$
    \State $\widehat{G} \gets \mathbf{1}\!\left[U_{\phi}(G_S) > 0.5\right]$ \Comment{binarize predicted occupancy probability}
    \State $\widehat{M} \gets T\bigl(c,\Call{OccupiedCoordinates}{\widehat{G}}\bigr)$
    \State \Return $S,\widehat{M}$
  \end{algorithmic}
\end{algorithm}

\subsection{Serialized Output} \label{app:complete_octllm_output}

Rendered meshes conceal what the language model itself produces. Below we therefore give one truncated image-to-3D response from OctLLM. It begins with a generated natural-language preamble and then contains 1,023 occupancy-byte tokens delimited by \texttt{<mesh\_bos>} and \texttt{<mesh\_eos>}. Whitespace is inserted only at token boundaries for typesetting. Note that the mesh tokens appear here in their literal vocabulary form, \texttt{<mesh0>} through \texttt{<mesh255>}, whereas the figures in this paper abbreviate the same tokens as \texttt{O0} through \texttt{O255} to keep the illustrations compact: \texttt{<mesh238>} in the output below and \texttt{O238} in \cref{fig:position_mask_design} denote the identical occupancy byte.

\begin{Verbatim}
I've produced a 3D mesh asset based on the image:
<mesh_bos> <mesh0> <mesh0> <mesh0> <mesh0> <mesh0> <mesh95> <mesh0> <mesh95> <mesh0>
<mesh0> <mesh0> <mesh0> <mesh175> <mesh0> <mesh175> <mesh0> <mesh0> <mesh0> <mesh0>
<mesh0> <mesh0> <mesh95> <mesh0> <mesh77> <mesh0> <mesh0> <mesh0> <mesh0> <mesh175>
<mesh0> <mesh142> <mesh0> <mesh0> <mesh245> <mesh0> <mesh245> <mesh0> <mesh0> <mesh0>

......

<mesh240> <mesh15> <mesh95> <mesh15> <mesh95> <mesh240> <mesh95> <mesh240> <mesh23> <mesh238>
<mesh23> <mesh250> <mesh31> <mesh128> <mesh254> <mesh128> <mesh85> <mesh234> <mesh1> <mesh87>
<mesh136> <mesh126> <mesh95> <mesh160> <mesh240> <mesh128> <mesh160> <mesh240> <mesh240> <mesh15>
<mesh95> <mesh15> <mesh126> <mesh248> <mesh_eos>
\end{Verbatim}

\section{Experimental Setup and Evaluation Protocol} \label{app:protocol}

\subsection{Additional Dataset Details} \label{app:data_construction}

\paragraph{Geometry preprocessing.}
The ShapeNet portion contains the airplane, car, chair, rifle, and table categories. Each mesh in the training dataset is centered and isotropically scaled to $[-1,1]^3$, and 100K surface points with face normals are used to build a depth-6 O-CNN~\cite{Wang2017} octree that is complete through depth 3. We discard assets that fail preprocessing or yield fewer than 100 or more than 4,000 mesh-byte tokens.

\paragraph{Caption annotation.}
We caption render indices 014--017 with a locally deployed Qwen3.5-9B~\cite{qwen35}. Because the S-Octree contains geometry but no color or material channels, we exclude appearance attributes that cannot be inferred from the 3D input. The prompt is:

\begin{PromptBox}
Describe this 3D object in English using only its geometry and shape.
Ignore color, material, texture, lighting, and background.
Focus on overall structure, major parts, proportions, and spatial
arrangement. Keep the description moderately detailed and concise.
Directly output the description without any other text. For example:
"A spider with multiple legs and a segmented body."
The description should not be longer than 100 words.
\end{PromptBox}

Caption decoding uses temperature $0.2$ and at most 200 new tokens, with thinking disabled. A fixed seed of 42 controls instruction-template and image-condition sampling; missing captions, renders, or mesh sequences are excluded.

\paragraph{Instruction data.}
We construct instruction-tuning conversations from 25 dialogue templates, following a protocol similar to ShapeLLM-Omni~\cite{shapellmomni}. For each sample we draw one template at random and fill it with the caption and the serialized S-Octree.

\subsection{Optimization Details} \label{app:optimization}

\begin{table}[H]
  \caption{Optimization settings not specified in \cref{sec:experiments}.}
  \label{tab:training_configuration}
  \tablestyle{4pt}{1.08}
  \begin{tabular}{@{}p{0.25\linewidth}p{0.69\linewidth}@{}}
    \toprule
    Setting & Value \\
    \midrule
    Input limits & 8,192 tokens and 262,144 image pixels \\
    OctLLM schedule & AdamW, peak learning rate $1\times10^{-4}$, cosine decay, and $10\%$ linear warmup \\
    Batch and precision & 2 samples/GPU, 2 gradient-accumulation steps, and bfloat16 \\
    Validation & $0.1\%$ of the training mixture \\
    Completion schedule & AdamW with cosine decay from $1\times10^{-4}$ to $1\times10^{-6}$ \\
    \bottomrule
  \end{tabular}
\end{table}

\subsection{Rendering Protocol} \label{app:rendering}

Meshes are centered and scaled to a unit bounding box, then rendered at $1024\times1024$ with Blender 4.0 and EEVEE. The 24 camera directions follow a spherical Hammersley sequence; camera radii range from 1.51 to 9.94 and horizontal fields of view from $7.5^\circ$ to $52.5^\circ$. We use identical lighting and camera sampling for all methods. For baselines without native textures, Hunyuan3D-2.1~\cite{hunyuan3d21} generates the material before rendering, and the resulting GLB materials are preserved.

\subsection{Generation Metrics} \label{app:generation_metrics}

\paragraph{Distributional metrics.}
We use render indices 014--017, yielding 4K images per complete distribution. For features $X_r$ and $X_g$ with empirical means $\mu_r,\mu_g$ and covariances $\Sigma_r,\Sigma_g$, FID is
\begin{equation}
  \operatorname{FID}=\lVert\mu_r-\mu_g\rVert_2^2+
  \operatorname{Tr}\!\left(\Sigma_r+\Sigma_g-2(\Sigma_r\Sigma_g)^{1/2}\right).
\end{equation}
KID is the unbiased squared MMD,
\begin{equation}
  \operatorname{KID}=
  \frac{\sum_{i\ne j}k(x_i,x_j)}{m(m-1)}+
  \frac{\sum_{i\ne j}k(y_i,y_j)}{n(n-1)}-
  \frac{2\sum_{i,j}k(x_i,y_j)}{mn},
  \quad k(x,y)=\left(\frac{x^\top y}{d}+1\right)^3.
\end{equation}
We compute both metrics with Inception-V3~\cite{szegedy2016inception} and DINOv2 ViT-L/14 register-token features~\cite{oquab2024dinov2}. DINOv2 KID averages 100 subsets of at most 1,000 samples, drawn without replacement with seed 42. KID values are scaled by 100 in \cref{tab:3d_generation}.

\paragraph{Condition alignment.}
CLIP uses ViT-L/14 embeddings~\cite{Radford2021} and reports $100\max(s,0)$ for cosine similarity $s$. Text-to-3D averages the caption similarity over views 014--017 for each asset. For image-to-3D, we compute for each asset the maximum similarity between the condition image and its 24 generated views, and then average this per-asset maximum over all assets. The maximum is used deliberately rather than a mean over views: the condition image shows the object from one unknown viewpoint, so only the rendered view whose pose best matches it provides a meaningful comparison, whereas averaging over all 24 views would penalize every method for viewpoints the condition never shows rather than for structural misalignment. The same protocol is applied to all methods.

\subsection{Understanding Evaluation} \label{app:understanding_metrics}

Each method receives its native representation: point clouds for PointLLM and ShapeLLM, OBJ text for LLaMA-Mesh, VQVAE indices for ShapeLLM-Omni, and S-Octrees for OctLLM. BLEU uses sentence-level method-1 smoothing, ROUGE reports F1, and Sentence-BERT and SimCSE use \texttt{all-mpnet-base-v2} and \texttt{princeton-nlp/sup-simcse-roberta-large}, respectively. Metrics are macro-averaged over matched assets and scaled by 100. Judge scores are averaged over valid outputs: 200/200 for all entries except LLaMA-Mesh (198/200 for both judges) and PointLLM-7B under GPT-img (199/200).

\Cref{tab:3d_understanding_full} adds PointLLM-13B and lexical metrics. Consistent with PointLLM~\cite{pointllm}, lexical overlap can favor short generic captions; our conclusions therefore rely on semantic similarity and the two judges.

\begin{table}[H]
  \caption{Complete PointLLM-200 metrics. All values are scaled by 100; the \colorbox{bestcolor}{best} result within each group is highlighted.}
  \label{tab:3d_understanding_full}
  \tablestyle{3pt}{1.15}
  \resizebox{\linewidth}{!}{
  \begin{tabular}{@{}l cccc ccc c cc cc@{}}
    \toprule
    & \multicolumn{8}{c}{$n$-gram overlap} & \multicolumn{2}{c}{Semantic} & \multicolumn{2}{c}{Judge} \\
    \cmidrule(lr){2-9} \cmidrule(lr){10-11} \cmidrule(lr){12-13}
    Method & BLEU-1 & BLEU-2 & BLEU-3 & BLEU-4 & ROUGE-1 & ROUGE-2 & ROUGE-L & METEOR & S-BERT & SimCSE & GPT-ref & GPT-img \\
    \midrule
    \multicolumn{13}{@{}l}{\textit{Task-specific 3D understanding models}} \\
    PointLLM-7B~\cite{pointllm}  & 5.43 & 1.59 & 0.74 & 0.48 & 10.46 & 0.89 & 7.67 & 13.47 & 48.05 & 48.63 & 46.06 & 45.63 \\
    PointLLM-13B~\cite{pointllm} & 5.45 & 1.63 & 0.75 & 0.48 & 10.46 & 0.96 & 7.64 & \best{13.61} & \best{48.85} & \best{49.78} & \best{48.09} & \best{48.06} \\
    ShapeLLM-7B~\cite{shapellm}  & \best{9.29} & \best{2.86} & \best{1.57} & \best{1.15} & \best{12.83} & \best{1.04} & \best{11.09} & 11.71 & 37.27 & 37.10 & 15.92 & 30.23 \\
    \cmidrule(lr){1-13}
    \multicolumn{13}{@{}l}{\textit{Multimodal LLMs}} \\
    LLaMA-Mesh~\cite{Wang2024a}  & \best{14.69} & \best{5.25} & \best{3.03} & \best{2.31} & \best{19.27} & \best{2.51} & \best{17.29} & \best{13.13} & 30.44 & 31.08 & 15.60 & 17.16 \\
    ShapeLLM-Omni~\cite{shapellmomni} & 4.57 & 1.24 & 0.58 & 0.41 & 8.30 & 0.56 & 6.41 & 11.53 & 36.08 & 34.28 & 22.27 & 18.14 \\
    \textbf{OctLLM (Ours)}       & 4.77 & 1.48 & 0.65 & 0.43 & 8.63 & 0.73 & 7.01 & 11.24 & \best{38.80} & \best{38.48} & \best{26.61} & \best{46.88} \\
    \bottomrule
  \end{tabular}}
\end{table}

\begin{figure}[H]
  \centering
  \includegraphics[width=0.95\linewidth,trim=12 22 16 6,clip]{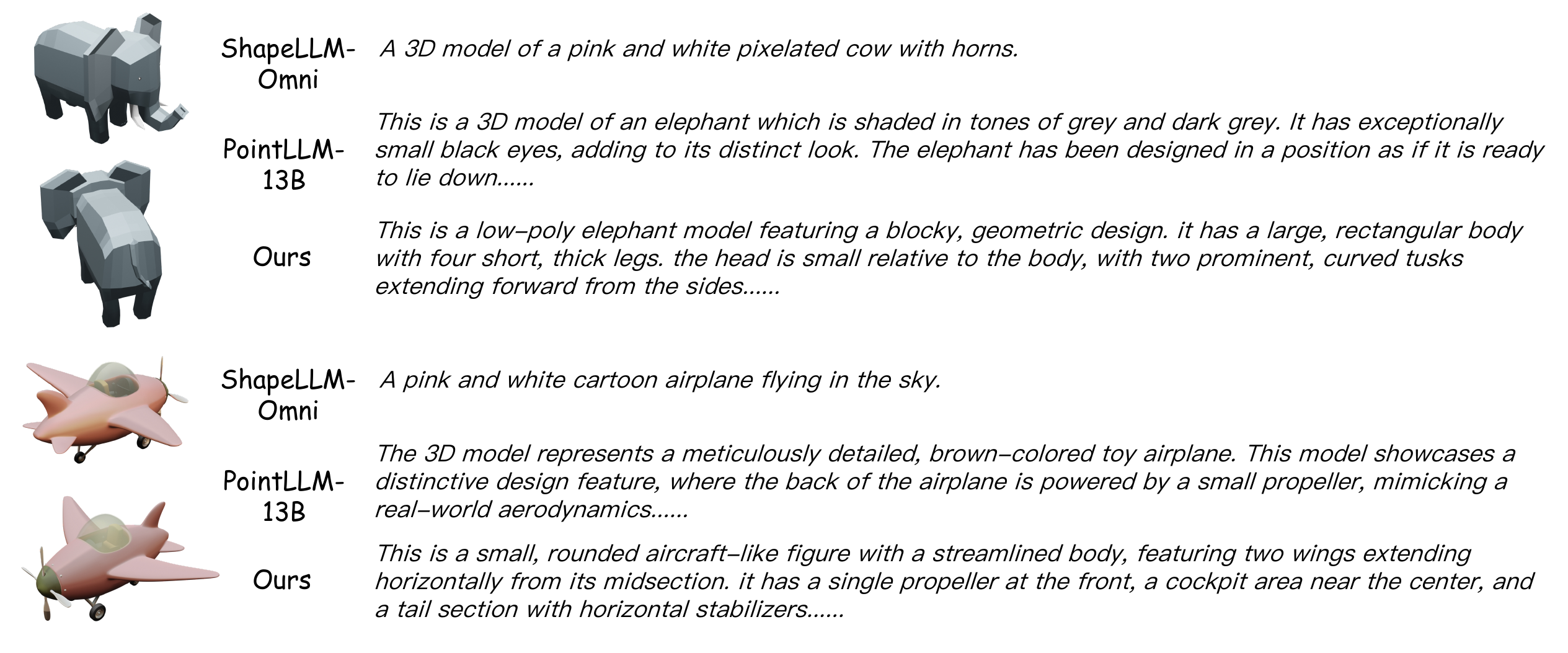}
  \caption{\textbf{3D understanding.} Descriptions of two held-out assets, each shown from two viewpoints and truncated for space. ShapeLLM-Omni produces terse descriptions and sometimes misses the object category, while PointLLM-13B emphasizes appearance over structure. OctLLM gives more complete descriptions of the visible parts and their spatial arrangement; \cref{app:complete_qualitative_text} reports its complete outputs.}
  \label{fig:qualitative_understanding}
\end{figure}

\section{Additional Results and Evaluation Details} \label{app:additional_generation}

\begin{figure}[p]
  \centering
  \includegraphics[width=\linewidth,keepaspectratio]{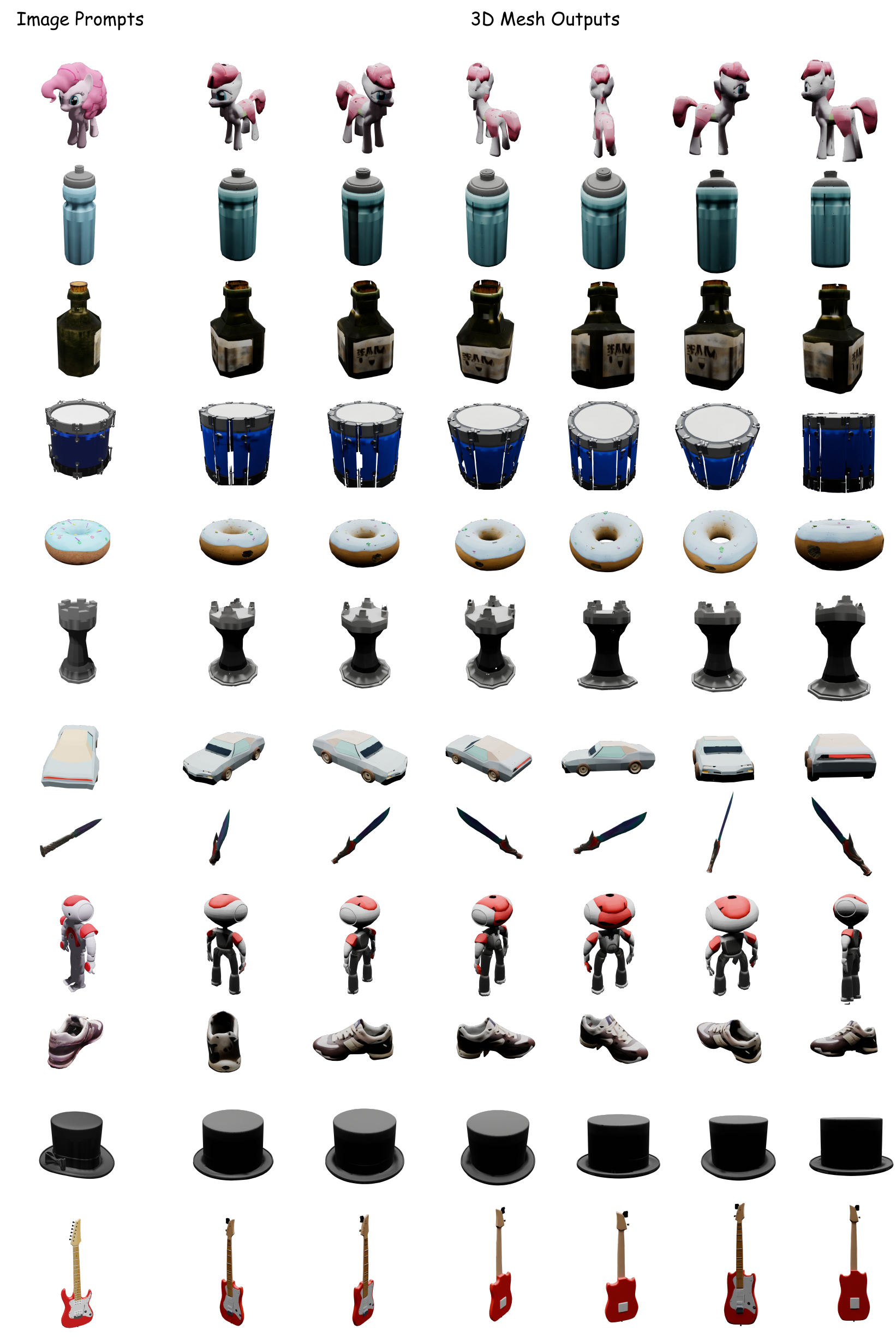}
  \caption{\textbf{Additional image-to-3D generation results.} Each row shows one image condition followed by the generated mesh from six viewpoints.}
  \label{fig:additional_image_generation}
\end{figure}

\begin{figure}[p]
  \centering
  \includegraphics[width=\linewidth,keepaspectratio]{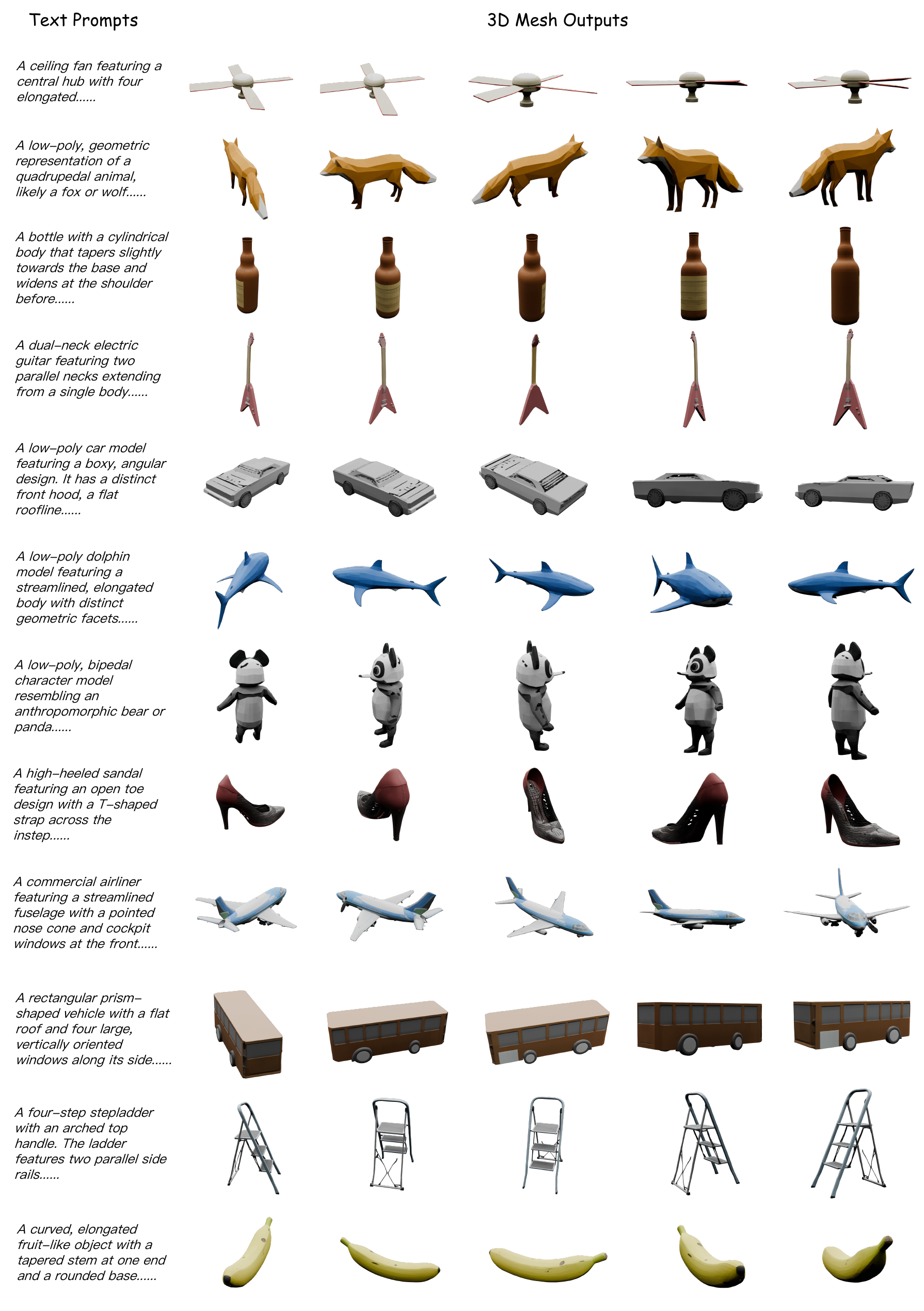}
  \caption{\textbf{Additional text-to-3D generation results.} Each row pairs a text condition, truncated in the figure, with five viewpoints of the generated mesh.}
  \label{fig:additional_text_generation}
\end{figure}

We provide additional qualitative results for OctLLM under image and text conditions. \Cref{fig:additional_image_generation,fig:additional_text_generation} show the generated meshes from multiple viewpoints.

\subsection{Complete Qualitative Conditions and Outputs} \label{app:complete_qualitative_text}

\paragraph{Text-to-3D conditions.}
At inference, each condition in \cref{fig:qualitative_text} follows the fixed instruction, ``Generate a 3D mesh based on the following text description:''; below we reproduce the complete condition strings in the same top-to-bottom order.

\begin{ExampleBox}
  \begin{enumerate}
    \item \emph{Biplane.} A biplane featuring two sets of wings stacked vertically, connected by struts and wires. The fuselage is elongated with a rounded nose housing a propeller. Tail assembly includes a vertical stabilizer and horizontal stabilizers. Landing gear consists of two wheels positioned under the front section. Overall design emphasizes symmetry and balanced proportions typical of early aviation aircraft.
    \item \emph{Stringed instrument.} A stringed musical instrument featuring a hollow, curved body with a central sound hole. It has a long, slender neck extending upwards from the body, terminating in a headstock equipped with tuning pegs. Six strings run along the length of the neck and across the body, anchored at the bridge. The overall form is symmetrical along its vertical axis, with smooth, rounded contours defining the body's shape.
    \item \emph{Quadrupedal dinosaur.} A quadrupedal dinosaur model featuring a robust body, four sturdy legs, and a long tail. The head is adorned with prominent horns and a frill, while the limbs are positioned to support its weight. The overall structure suggests a powerful stance, typical of herbivorous dinosaurs.
    \item \emph{Pedestal sink.} A pedestal sink featuring a rectangular basin with rounded corners, supported by a cylindrical column. The faucet is centrally mounted on the back edge of the basin, extending forward with a curved spout. The overall design is compact and functional, with smooth surfaces and clean lines.
  \end{enumerate}
\end{ExampleBox}

\paragraph{3D-understanding outputs.}
The two OctLLM descriptions in \cref{fig:qualitative_understanding} were shortened only in the figure. Their complete, unedited outputs are:\nopagebreak

\begin{ExampleBox}
  \emph{Elephant (top).} This is a low-poly elephant model featuring a blocky, geometric design. it has a large, rectangular body with four short, thick legs. the head is small relative to the body, with two prominent, curved tusks extending forward from the sides. large, fan-shaped ears are positioned on either side of the head. the trunk hangs down between the legs. the overall shape is angular and simplified, lacking smooth curves

  \emph{Aircraft (bottom).} This is a small, rounded aircraft-like figure with a streamlined body, featuring two wings extending horizontally from its midsection. it has a single propeller at the front, a cockpit area near the center, and a tail section with horizontal stabilizers. the design is compact and aerodynamic, with smooth curves and a simple, geometric form
\end{ExampleBox}

\subsection{Judge Prompts} \label{app:judge_prompts}

The reference-based judge receives the human caption and the model caption and returns a score with a short justification:

\begin{PromptBox}
Evaluate a model-generated caption against a human-generated caption
(ground truth) for a 3D model. Identify the aspects mentioned in the
human caption and calculate the percentage of these aspects correctly
mentioned or partially matched in the model caption. Score from 0 to
100, where each aspect contributes equally to the score. Consider
similar concepts for partial score.

Provide your score (0-100) and a short justification (less than 15
words) in the format of 'score#reason'

Example:
Human: A white brown skeleton
Model: This is a 3D model of a small, cartoon-like robot. It has a
spherical body and is covered in a layer of white dust.
Output: 50#mention white; skeleton and robot have similar appearance.

Now score the following:
Human: {ground_truth}
Model: {model_output}
Output:
\end{PromptBox}

The render-grounded judge receives four views of the asset and the model caption, but not the reference:

\begin{PromptBox}
Render-based GPT judge for PointLLM-200 mesh captioning.

Inputs:
- Four RGB renders of the same 3D object: front, right, back, and left.
- One model-generated caption.
- The human caption is intentionally not provided.

Goal:
Score how faithfully the caption describes the visible 3D object in the
renders. The score is a scalar integer from 0 to 100, where higher is
better.

Rubric:
1. Core object identity and function (0-35): award high credit when the
   caption names the correct object category or a close synonym.
2. Geometry, structure, and parts (0-25): reward correct major visible
   components, shape, attachments, symmetry, and distinctive geometry.
3. Color, material, and texture (0-15): reward correct visible colors,
   material cues, texture patterns, and surface finish.
4. Fine-grained attributes and style (0-15): reward accurate style,
   decorative motifs, proportions, pose, orientation, and special
   identifying features.
5. Caption quality and specificity (0-10): reward concise but
   informative captions and penalize vague or repetitive captions.

Penalties and caps:
- Severe hallucination should reduce the score.
- Empty, non-natural-language, or token-like captions should score 0-10.
- Captions for a completely different object should score 0-20.
- Broadly correct category with many wrong details should usually score
  40-70.

Output strict JSON only:
{"score": <integer 0-100>, "reason": "<one short sentence>",
 "matched": ["..."], "errors": ["..."]}
\end{PromptBox}

\subsection{Autoregressive Inference Efficiency} \label{app:inference_efficiency}

We compare text-to-3D token counts and autoregressive latency on a fixed set of 200 held-out Toys4K prompts. Each model runs sequentially at batch size one on a single NVIDIA GeForce RTX 5090, with one complete warm-up generation and its recommended decoding settings. We report means over all 200 outputs. Latency is synchronized CUDA-event time over each method's autoregressive generation region; input processing, downstream 3D reconstruction, and rendering are excluded. For OctLLM, the timed region includes the complete constrained octree loop and the per-step structural checks needed to select the next query.

\begin{table}[H]
  \caption{Text-to-3D token count and autoregressive latency on a fixed 200-prompt Toys4K subset. Values are means on one RTX 5090. Token units are method-native and differ across representations.}
  \label{tab:inference_efficiency}
  \tablestyle{6pt}{1.15}
  \begin{tabular}{@{}l rrr@{}}
    \toprule
    Method & Output tokens & Structure tokens & AR latency (s)~$\downarrow$ \\
    \midrule
    ShapeLLM-Omni~\cite{shapellmomni} & 1,026.00 & 1,024.00 & 19.92 \\
    AR3D-R1~\cite{rl3dr12025}         & 1,026.00 & 1,024.00 & 23.50 \\
    LLaMA-Mesh~\cite{Wang2024a}       & 4,443.50 & 4,442.53 & 85.30 \\
    OctGPT~\cite{Wei2025}             & 10,033.44 & 2,757.48 & 52.53 \\
    \textbf{OctLLM (Ours)}            & 1,277.06 & 1,262.06 & 37.59 \\
    \bottomrule
  \end{tabular}
\end{table}

Output tokens follow each method's native output unit, whereas structure tokens report its representation-specific structural sequence: VQ indices for ShapeLLM-Omni and AR3D-R1, OBJ BPE tokens for LLaMA-Mesh, split-node positions for OctGPT, and occupancy-byte tokens for OctLLM. These counts characterize representation length rather than a shared vocabulary. OctLLM averages 1,277.06 output tokens, only modestly more than the fixed-token ShapeLLM-Omni and AR3D-R1 representations and substantially fewer than LLaMA-Mesh and OctGPT. Its optimized constrained decoder takes 37.59 seconds on average, making its latency competitive among the compared methods and faster than OctGPT and LLaMA-Mesh.

\section{Ablation Study} \label{app:ablation}

All variants are trained on ShapeNet airplanes~\cite{Chang2015} under a fixed data and optimization protocol, using four 80\,GB A100 GPUs. Training lasts at most 15 epochs, and we report the checkpoint with the best performance on a 5\% validation split. Unless stated otherwise, they route 8 branches through layers $\{0,4,8,12,16,20,24,27\}$ for training efficiency. We compute Inception-V3 FID/KID on rendered normal maps and Sentence-BERT/SimCSE similarity for understanding; KID is scaled by 100. These subset results are not directly comparable with the full-model Toys4K evaluation.

\paragraph{Explicit 3D Language versus Latent Codes.}
We replace the S-Octree with the 1,024 VQVAE tokens used by ShapeLLM-Omni~\cite{shapellmomni}, keeping all other settings fixed. In \cref{tab:representation_ablation}, the S-Octree reduces FID by 20.5\% and improves both understanding metrics at the same KID. Its coordinate-anchored occupancy tokens therefore benefit both generation and understanding.
Because the representation is the only changed component, this result directly supports our central premise: exposing spatial coordinates and hierarchy to the LLM is more effective than asking it to operate on opaque latent indices.

\begin{table}[H]
  \caption{Ablation on the 3D representation. All settings except the representation are fixed; KID is scaled by 100. Generation metrics are computed on the image-to-3D task.}
  \label{tab:representation_ablation}
  \tablestyle{6pt}{1.15}
  \begin{tabular}{l cc cc}
    \toprule
    & \multicolumn{2}{c}{Generation} & \multicolumn{2}{c}{Understanding} \\
    \cmidrule(lr){2-3} \cmidrule(lr){4-5}
    Representation & FID~$\downarrow$ & KID~$\downarrow$ & Sentence-BERT~$\uparrow$ & SimCSE~$\uparrow$ \\
    \midrule
    VQVAE tokens (1,024) & 33.10 & \best{0.80} & 67.15 & 68.87 \\
    S-Octree (Ours)      & \best{26.31} & \best{0.80} & \best{69.47} & \best{70.28} \\
    \bottomrule
  \end{tabular}
\end{table}

\paragraph{Preserving Language Ability.}
We compare OctLLM with a rank-64 LoRA control~\cite{lora} that also mixes UltraChat~\cite{ultrachat} into 3D training, following ShapeLLM-Omni~\cite{shapellmomni}. Despite the additional text data, LoRA loses 5.1 MMLU and 18.1 HellaSwag points relative to the backbone (\cref{tab:language_preservation_ablation}). OctLLM matches the backbone on MMLU, HellaSwag, and IFEval and remains within 0.5 points on GSM8K because 3D gradients update only the routed branches, not the shared language pathway.
Together with the representation ablation, this isolates the two complementary insights of OctLLM: explicit spatial tokens improve 3D modeling, while parameter-isolated routing adds this capability without overwriting the pretrained language pathway.

\begin{table}[H]
  \caption{Language preservation after ShapeNet-airplane training. The LoRA control includes UltraChat; the frozen backbone is shown as a gray reference.}
  \label{tab:language_preservation_ablation}
  \tablestyle{5pt}{1.15}
  \begin{tabular}{l cccc}
    \toprule
    Method & MMLU~$\uparrow$ & HellaSwag~$\uparrow$ & GSM8K~$\uparrow$ & IFEval~$\uparrow$ \\
    \midrule
    \rowcolor{graycolor}
    Frozen Qwen2.5-VL-7B-Instruct (reference)
      & 64.65 & 67.44 & 83.54 & 75.41 \\
    LoRA + UltraChat
      & 59.58 & 49.34 & 69.97 & 59.33 \\
    OctLLM (Ours)
      & \best{64.65} & \best{67.44} & \best{83.09} & \best{75.41} \\
    \bottomrule
  \end{tabular}
\end{table}

\paragraph{Components of the S-Octree Pathway.}
\Cref{tab:component_ablation} isolates the three S-Octree components. Removing predictive mask tokens causes the largest degradation; 3D position embeddings and mask suppression provide additional gains.
This pattern supports their complementary roles: mask queries specify where occupancy is predicted, position embeddings ground each query in 3D, and mask suppression keeps previous query placeholders from contaminating later context. The full model consequently combines spatially grounded prediction with a clean autoregressive history.

\begin{table}[H]
  \caption{S-Octree component ablations for image-to-3D generation. KID is scaled by 100.}
  \label{tab:component_ablation}
  \tablestyle{6pt}{1.15}
  \begin{tabular}{l cc}
    \toprule
    Method & FID~$\downarrow$ & KID~$\downarrow$ \\
    \midrule
    w/o 3D Position Embedding & 32.84 & 1.00 \\
    w/o Mask Token            & 61.08 & 3.87 \\
    w/o Mask Suppression      & 28.03 & 0.94 \\
    Full model                & \best{26.31} & \best{0.80} \\
    \bottomrule
  \end{tabular}
\end{table}

\paragraph{Placement of Routed 3D Branches.}
With eight routed branches, uniform placement gives the best generation scores under both conditions (\cref{tab:branch_layer_ablation}), while the first- and last-layer placements favor understanding. The last eight layers also outperform the first eight on every generation metric. Although uniform placement does not maximize understanding, it remains within 4.2 points of the best targeted placement on both semantic metrics, while reducing image- and text-conditioned FID by 22.7\% and 17.8\% relative to the next-best placement. We therefore select uniform depth coverage as the stronger joint-task trade-off and assign the full model's four additional branches to later layers.

\begin{table}[H]
  \caption{Placement of eight routed 3D branches. FID/KID (scaled by 100) evaluate generation; Sentence-BERT and SimCSE evaluate understanding.}
  \label{tab:branch_layer_ablation}
  \tablestyle{5pt}{1.15}
  \resizebox{\linewidth}{!}{
  \begin{tabular}{l cccc cc}
    \toprule
    & \multicolumn{4}{c}{Generation} & \multicolumn{2}{c}{Understanding} \\
    \cmidrule(lr){2-5} \cmidrule(lr){6-7}
    Placement & FID (image)~$\downarrow$ & KID (image)~$\downarrow$ & FID (text)~$\downarrow$ & KID (text)~$\downarrow$ & Sentence-BERT~$\uparrow$ & SimCSE~$\uparrow$ \\
    \midrule
    First 8/28 layers          & 36.63 & 1.70 & 38.97 & 1.62 & 72.26 & \best{74.37} \\
    Last 8/28 layers           & 34.03 & 1.31 & 37.09 & 1.49 & \best{73.65} & 73.71 \\
    Uniform 8/28 layers (Ours) & \best{26.31} & \best{0.80} & \best{30.48} & \best{1.07} & 69.47 & 70.28 \\
    \bottomrule
  \end{tabular}}
\end{table}

\noindent
\begin{minipage}[t]{0.565\linewidth}
  \vspace{0pt}
  \textbf{Scaling Routed 3D Capacity.}
  The 4-, 6-, and 8-branch variants use layers $\{0,8,16,24\}$, $\{0,8,12,16,24,27\}$, and $\{0,4,8,12,16,20,24,27\}$, respectively, corresponding to 0.93B, 1.40B, and 1.86B trainable parameters.

  In \cref{fig:branch_scaling}, FID changes little from four to six branches but falls from 37.63 to 26.31 with eight; KID likewise improves to 0.80. The gain is therefore not proportional to the budget at the smallest sizes, but the controlled eight-branch result shows that allocating sufficient parameters across the full depth substantially improves generation quality. This supports scaling the isolated 3D pathway rather than modifying the shared backbone when more 3D capacity is needed.
\end{minipage}\hfill
\begin{minipage}[t]{0.405\linewidth}
  \vspace{0pt}
  \centering
  \includegraphics[width=\linewidth]{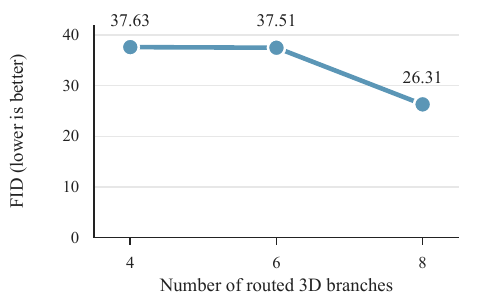}
  \captionsetup{hypcap=false}
  \captionof{figure}{\textbf{Scaling routed 3D capacity.} Image-to-3D FID on ShapeNet airplanes. Only the branch count changes.}
  \label{fig:branch_scaling}
\end{minipage}

\paragraph{Full-Rank 3D Branches versus Token-Routed LoRA.}
The rank-64 LoRA control in \cref{tab:language_preservation_ablation} tests language forgetting, but leaves open whether token-routed LoRA can preserve language while providing sufficient 3D capacity. We therefore replace the default eight full-rank 3D branches with rank-256 LoRA adapters in all 28 layers and route only 3D tokens through them, while keeping the ShapeNet-airplane dataset protocol fixed. The added 3D vocabulary requires trainable embedding and language-head rows, so a row-wise stop-gradient restricts updates to the new token rows. This control matches the backbone on MMLU and HellaSwag, as does the dual-stream model, but the full-rank branches perform better on three of four generation metrics (\cref{tab:token_routed_lora}): image-conditioned FID falls from 32.44 to 26.31 and KID from 1.06 to 0.80, while text-conditioned FID improves from 30.93 to 30.48. On the full training data we further observed that the full-rank branches train stably, whereas high-rank LoRA suffers gradient explosions: a low-rank residual does not give 3D occupancy enough expressive room, and enlarging the rank makes large-scale training unstable. Token-routed LoRA therefore preserves language ability, but does not match the 3D capacity or optimization stability of dedicated full-rank branches.

\begin{table}[H]
  \caption{Full-rank dual-stream branches versus token-routed LoRA on ShapeNet airplanes. FID/KID evaluate image- and text-conditioned 3D generation (KID scaled by 100); language scores are reference values because both variants match the frozen backbone.}
  \label{tab:token_routed_lora}
  \tablestyle{4pt}{1.15}
  \resizebox{\linewidth}{!}{
  \begin{tabular}{@{}l cccc cc@{}}
    \toprule
    & \multicolumn{4}{c}{3D Generation} & \multicolumn{2}{c}{Language Ability (reference)} \\
    \cmidrule(lr){2-5} \cmidrule(lr){6-7}
    Method & FID (image)~$\downarrow$ & KID (image)~$\downarrow$ & FID (text)~$\downarrow$ & KID (text)~$\downarrow$ & MMLU~$\uparrow$ & HellaSwag~$\uparrow$ \\
    \midrule
    Dual-stream branches (Ours) & \best{26.31} & \best{0.80} & \best{30.48} & 1.07 & 64.65 & 67.44 \\
    Token-routed LoRA (rank 256) & 32.44 & 1.06 & 30.93 & \best{0.98} & 64.65 & 67.44 \\
    \bottomrule
  \end{tabular}}
\end{table}

\begin{figure}[H]
  \centering
  \includegraphics[width=\linewidth]{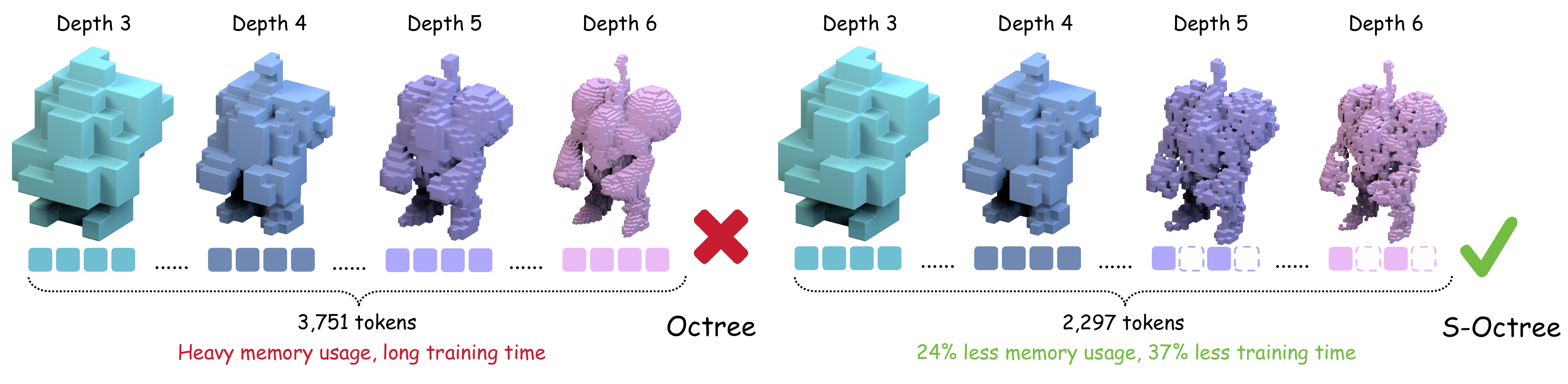}
  \caption{Full octree versus S-Octree on the same asset, shown at depths 3 to 6 with the corresponding mesh-token sequence beneath each level. Pruning part of the second-to-last level, and with it the finest-level descendants, shortens the sequence the LLM must write without altering the coarse structure it predicts.}
  \label{fig:soctree}
\end{figure}

\begin{figure}[H]
  \centering
  \begin{subfigure}[t]{0.48\linewidth}
    \centering
    \includegraphics[width=\linewidth]{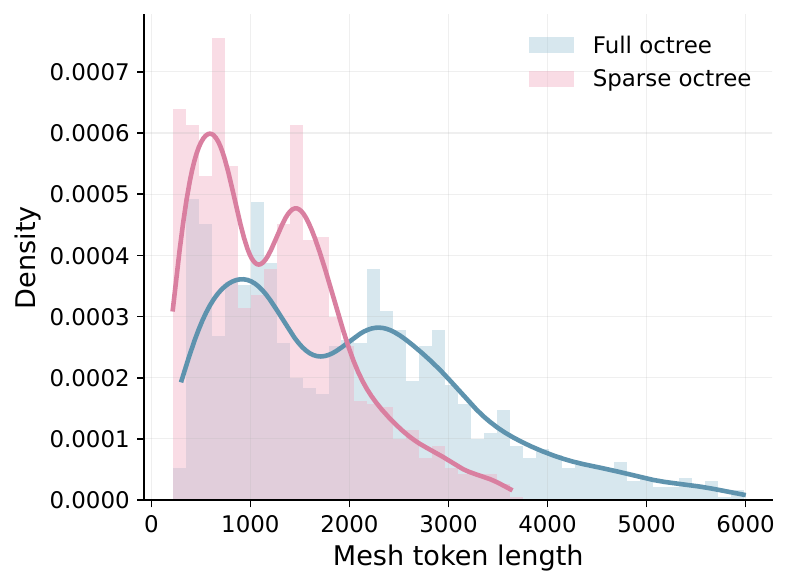}
    \caption{Mesh-token length distribution.}
    \label{fig:soctree_token_distribution}
  \end{subfigure}
  \hfill
  \begin{subfigure}[t]{0.48\linewidth}
    \centering
    \includegraphics[width=\linewidth]{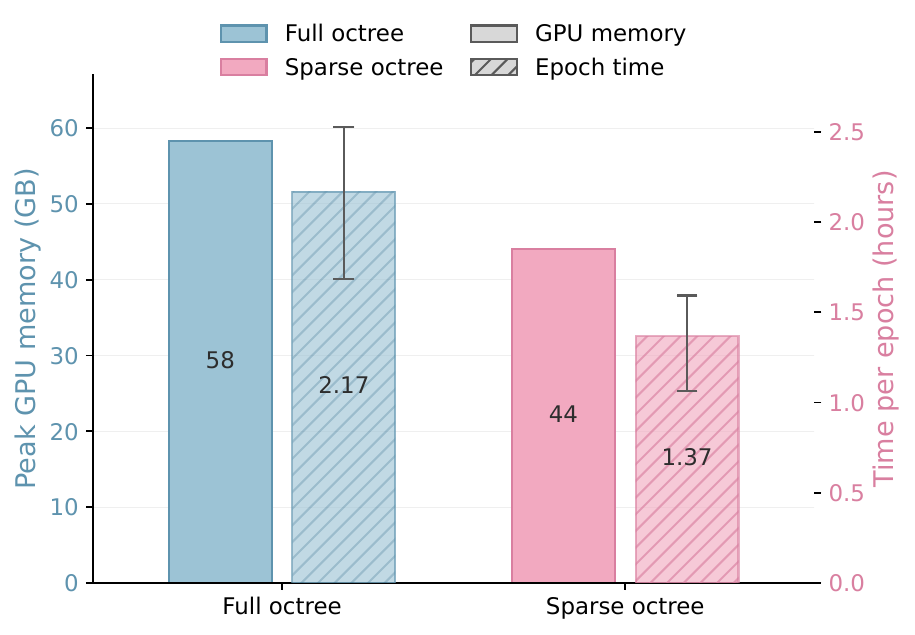}
    \caption{Training resource cost.}
    \label{fig:soctree_resource_cost}
  \end{subfigure}
  \caption{Practical benefits of the S-Octree representation on five selected ShapeNet categories. Compared with full octrees, S-Octrees shift the mesh-token length distribution toward shorter sequences and reduce both peak GPU memory and training time per epoch.}
  \label{fig:soctree_efficiency}
\end{figure}

\paragraph{S-Octree Efficiency.}
For the example in \cref{fig:soctree}, sparsification reduces the sequence from 3,751 to 2,297 mesh tokens while preserving the coarse structure. \Cref{fig:soctree_efficiency} evaluates the same effect over five ShapeNet categories.
Full-octree sequences extend to nearly 6K tokens. Measured on the same four 80\,GB A100 GPUs for both settings, S-Octree reduces peak memory from 58\,GB to 44\,GB and epoch time from 2.17 to 1.37 hours ($1.6\times$). The memory saving matters in practice, since it brings training within the 48\,GB budget of widely available accelerators such as the L40 instead of requiring 80\,GB cards. The completion network restores the pruned fine-level occupancy.

To assess generation quality, we additionally train on complete octree sequences without random emptying, using the same hardware, ShapeNet-airplane protocol, and eight-branch architecture. Under image conditioning, this model achieves FID 27.54 and KID 0.83 (scaled by 100), compared with 26.31 and 0.80 for S-Octree. The scores are close, with the full octree slightly worse. Thus, in this ablation, random emptying followed by completion preserves generation quality while reducing training time, memory usage, and the autoregressive decoding workload.

\paragraph{Sparse Voxel Completion Design.}
We ablate the loss and architecture of the 3D U-Net that completes the S-Octree occupancy before TRELLIS decoding. In \cref{tab:unet_ablation}, the default bottlenecked U-Net without cross-scale skip connections achieves the best FID and KID with binary cross-entropy. Replacing BCE with Dice loss slightly degrades both metrics, while adding skip connections or aligning incomplete-shape latents to a pretrained encoder causes substantially larger degradation. These results support binary occupancy supervision through a compact bottleneck for recovering missing fine structure, motivating the completion design used in OctLLM.

\begin{table}[H]
  \caption{Ablation on the sparse voxel completion network for image-to-3D generation. KID is scaled by 100.}
  \label{tab:unet_ablation}
  \tablestyle{6pt}{1.15}
  \begin{tabular}{l cc}
    \toprule
    Completion design & FID~$\downarrow$ & KID~$\downarrow$ \\
    \midrule
    Encoder alignment + MSE          & 58.93 & 3.49 \\
    3D U-Net + skip connections + BCE & 52.14 & 2.92 \\
    3D U-Net + Dice                   & 38.89 & 1.87 \\
    3D U-Net + BCE (Ours)          & \best{36.86} & \best{1.60} \\
    \bottomrule
  \end{tabular}
\end{table}

\section{Limitations} \label{app:limitations}

OctLLM has several limitations. First, its variable-length S-Octree allocates fewer tokens to simple shapes and more to complex ones, yet averages 1,793 tokens across our training set, compared with ShapeLLM-Omni's fixed 1,024-token representation. The longer sequences increase both training time and GPU memory usage. Spatially adaptive octree refinement may retain this flexibility with shorter sequences. Second, geometry passes through two learned predictions---S-Octree generation and 3D U-Net completion---so upstream occupancy errors can propagate or be amplified, producing holes and other artifacts that lower generation quality. Third, OctLLM does not support 3D understanding of appearance or 3D editing. Future work will add native color and material modeling, so OctLLM no longer depends on an external 3D generator and can describe appearance as well as geometry. We also plan to introduce 3D editing in subsequent versions of OctLLM.

\clearpage
\FloatBarrier

\end{document}